\documentclass[letterpaper, 10 pt, conference]{ieeeconf}
\IEEEoverridecommandlockouts
\useRomanappendicesfalse
\usepackage[utf8]{inputenc}

\usepackage[dvips]{graphicx}
\usepackage{xcolor}
\definecolor{mygreen}{rgb}{0,0.6,0}
\definecolor{mygray}{rgb}{0.5,0.5,0.5}
\definecolor{mymauve}{rgb}{0.58,0,0.82}

\usepackage{mathtools}
\usepackage{amsfonts}
\usepackage{amssymb}
\usepackage{multirow}
\usepackage{hyperref}
\usepackage{tabularx}
\usepackage{url}
\usepackage{bm}
\usepackage{enumerate}
\usepackage{subcaption}
\usepackage{svg}
\usepackage{mathtools}
\newcommand{\ttb}{\ttfamily\bfseries}
\usepackage{listings}
\lstdefinestyle{python_snippet}{
  backgroundcolor=\color{white},   % choose the background color; you must add \usepackage{color} or \usepackage{xcolor}; should come as last argument
  basicstyle=\footnotesize,        % the size of the fonts that are used for the code
  breakatwhitespace=true,         % sets if automatic breaks should only happen at whitespace
  breaklines=true,                 % sets automatic line breaking
  captionpos=b,                    % sets the caption-position to bottom
  commentstyle=\color{mygreen},    % comment style
  frame=single,	                   % adds a frame around the code
  keepspaces=true,                 % keeps spaces in text, useful for keeping indentation of code (possibly needs columns=flexible)
  keywordstyle=\color{blue},       % keyword style
  numbers=left,                    % where to put the line-numbers; possible values are (none, left, right)
  numbersep=5pt,                   % how far the line-numbers are from the code
  numberstyle=\tiny\color{mygray}, % the style that is used for the line-numbers
  rulecolor=\color{black},         % if not set, the frame-color may be changed on line-breaks within not-black text (e.g. comments (green here))
  showspaces=false,                % show spaces everywhere adding particular underscores; it overrides 'showstringspaces'
  showstringspaces=false,          % underline spaces within strings only
  showtabs=false,                  % show tabs within strings adding particular underscores
  stepnumber=2,                    % the step between two line-numbers. If it's 1, each line will be numbered
  stringstyle=\color{mymauve},     % string literal style
  tabsize=2,	                   % sets default tabsize to 2 spaces
}
\usepackage{stmaryrd}
\usepackage{soul}
\definecolor{ld_blue}{rgb}{0, 0, 0.6}
\definecolor{el_purple}{rgb}{0.6, 0, 0.6}
\definecolor{nm_orange}{rgb}{0.8, 0.4, 0.0}
\definecolor{jc_pink}{rgb}{0.6, 0.0, 0.3}

\definecolor{deepblue}{rgb}{0,0,0.5}
\definecolor{deepred}{rgb}{0.6,0,0}
\definecolor{deepgreen}{rgb}{0,0.5,0}
\definecolor{lightgray}{gray}{0.95} %the shade of grey that stack exchange uses

\newcommand{\kron}{\otimes}

\newcommand{\SE}[1]{\mathrm{SE}(#1)}

\newcommand{\matr}[1]{\boldsymbol{#1}}

\newcommand{\zeros}[2]{
\ifthenelse{\equal{#2}{1}}{\vect{0}_{#1}}{\matr{\cancel{O}}_{#1 \times #2}}
}
\newcommand{\ones}[2]{
\ifthenelse{\equal{#2}{1}}{\vect{1}_{#1}}{\matr{1}_{#1 \kron #2}}
}
\newcounter{simulationcase}

\newcommand{\F}{\mathcal{F}}

\newcommand{\old}[1]{}  %comment not showed

\newcommand{\real}[1]{\mathbb{R}^{#1}{}}

\newcommand{\cross}[1]{[#1]_{\times}\!}

\newcommand{\norm}[1]{\lVert#1\rVert}

\DeclareMathOperator{\Log}{Log}

\title{Optimal Control of Walkers with Parallel Actuation}
\author{Ludovic De Matteïs$^{1, 2, *}$, Virgile Batto$^{1, 3}$, Justin Carpentier$^{2}$, Nicolas Mansard$^{1, 4}$
\thanks{This work is suported by ROBOTEX 2.0 (ROBOTEX ANR-10-EQPX-0044 and TIRREX ANR-21-ESRE-0015), ANITI (ANR-19-P3IA-0004), by the French government (INEXACT ANR-22-CE33-0007 and "Investissements d'avenir" ANR-19-P3IA-0001) (PRAIRIE 3IA Institute), and by the Louis Vuitton ENS Chair on Artificial Intelligence}
\thanks{$^1$ Gepetto, LAAS-CNRS, Université de Toulouse, France}
\thanks{$^2$ Inria, École normale Supérieure, PSL Research University, Paris, France}
\thanks{$^3$ Auctus, Inria, centre de l'université de Bordeaux, Talence, France}
\thanks{$^4$ Artificial and Natural Intelligence Toulouse Institute, France}
\thanks{$^*$ Corresponding author: \href{mailto:ludovic.de-matteis@laas.fr}{ludovic.de-matteis@laas.fr}}
}
\date{July 2024}

\begin{document}

\maketitle

\begin{abstract}
Legged robots with closed-loop kinematic chains are increasingly prevalent due to their increased mobility and efficiency. 
Yet, most motion generation methods rely on serial-chain approximations, sidestepping their specific constraints and dynamics.
This leads to suboptimal motions and limits the adaptability of these methods to diverse kinematic structures.
We propose a comprehensive motion generation method that explicitly incorporates closed-loop kinematics and their associated constraints in an optimal control problem, integrating kinematic closure conditions and their analytical derivatives. 
This allows the solver to leverage the non-linear transmission effects inherent to closed-chain mechanisms, reducing peak actuator efforts and expanding their effective operating range. 
Unlike previous methods, our framework does not require serial approximations, enabling more accurate and efficient motion strategies. We also are able to generate the motion of more complex robots for which an approximate serial chain does not exist. 
We validate our approach through simulations and experiments, demonstrating superior performance in complex tasks such as rapid locomotion and stair negotiation. This method enhances the capabilities of current closed-loop robots and broadens the design space for future kinematic architectures.
\end{abstract}
\section{Introduction} \label{sec:introduction}
Recent progress in biped locomotion result from the sound combination of more mature motion generation techniques and continuous improvements in robot design and hardware \cite{tadeuszmikolajczykRecentAdvancesBipedal2022}.
Several advancements have recently demonstrated the advantages of leveraging parallel kinematic chains to boost the dynamic capabilities of robots \cite{jean-pierremerletParallelRobotsSolid2006}.
This architecture offers benefits like lighter lower limbs and improved impact absorption \cite{virgilebattoComparativeMetricsAdvanced2023} at the cost of introducing more complex dynamics, \cite{carlosmastalliCrocoddylEfficientVersatile2019} eventually making the robot more difficult to simulate and control \cite{g.f.liuAnalysisControlRedundant2001}.
%%%%%%
\begin{figure}[t]
    \centering
    \hfill
    \begin{subfigure}{0.215\linewidth}
        \centering
        \includegraphics[trim={0cm 0cm 0cm 0cm}, clip, width=\linewidth]{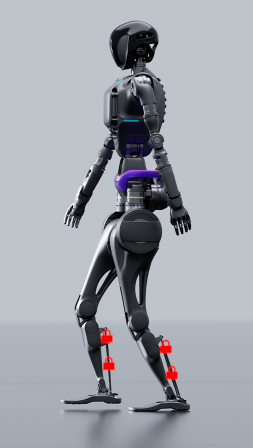}
    \end{subfigure}
    \hfill
    \begin{subfigure}{0.22\linewidth}
        \centering
        \includegraphics[trim={0cm 0cm 0cm 0cm}, clip, width=\linewidth]{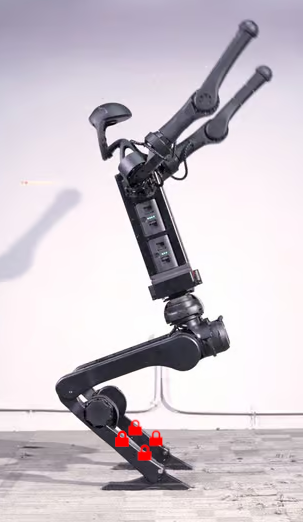}
    \end{subfigure}
    \hfill
    \begin{subfigure}{0.19\linewidth}
        \centering
        \includegraphics[trim={0cm 0cm 0cm 0cm}, clip, width=\linewidth]{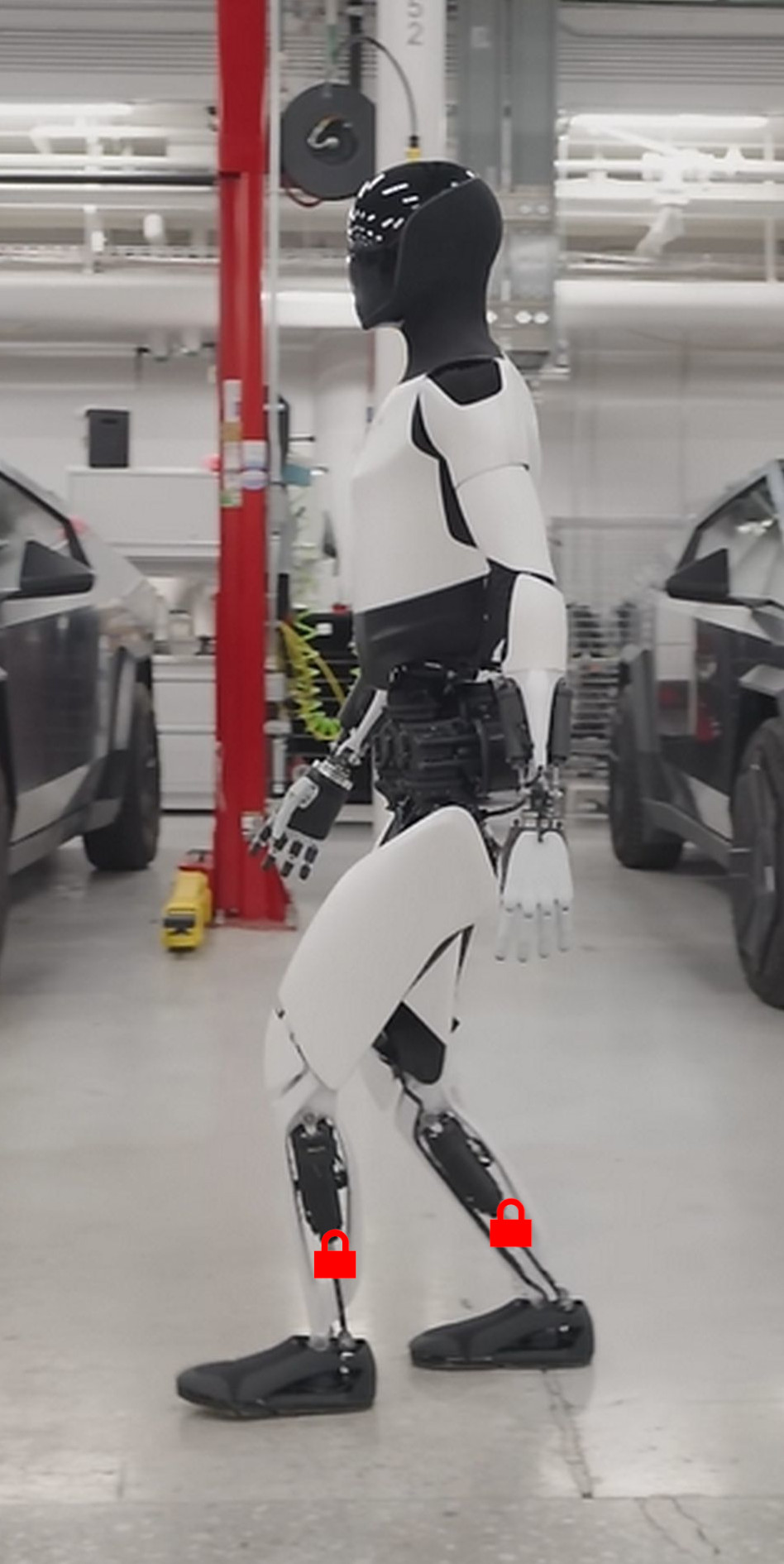}
    \end{subfigure}
    \hfill
    \begin{subfigure}{0.20\linewidth}
        \centering
        \includegraphics[trim={0cm 0cm 0cm 0cm}, clip, width=\linewidth]{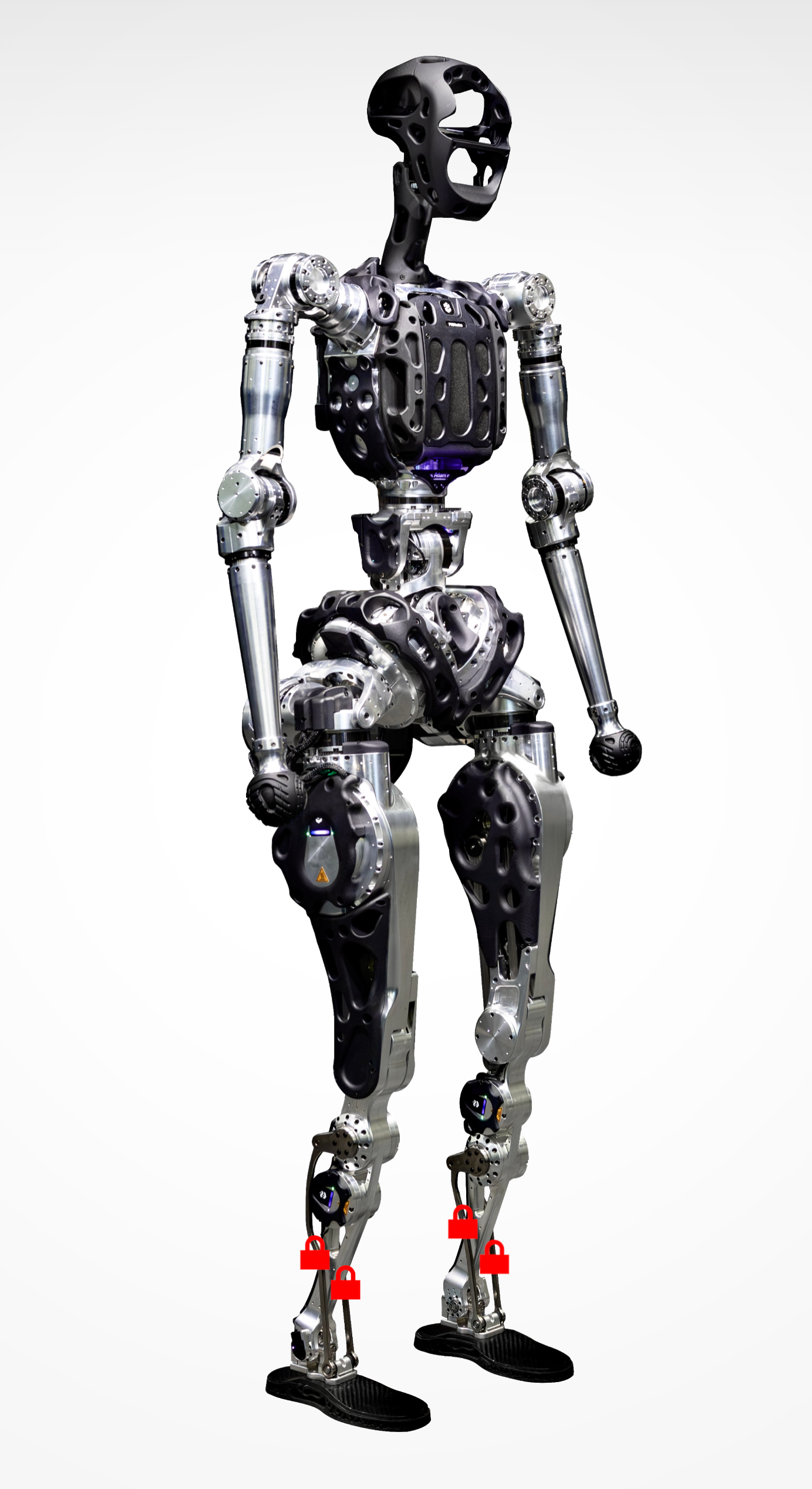}
    \end{subfigure}
    \hfill
    \\
    \hfill
    \begin{subfigure}{0.25\linewidth}
        \centering
        \includegraphics[trim={0cm 0cm 0cm 0cm}, clip, width=\linewidth]{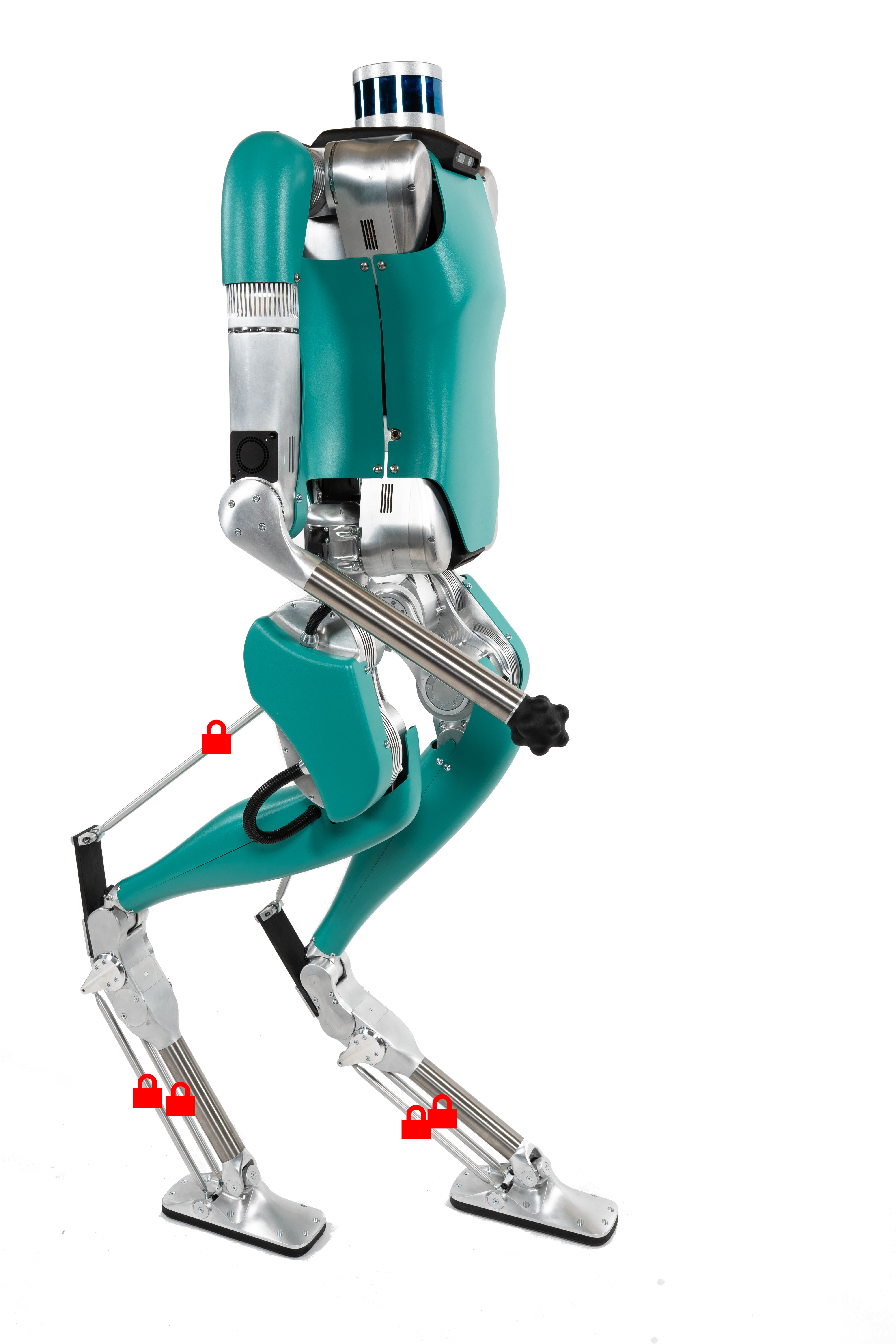}
    \end{subfigure}
    \hfill
    \begin{subfigure}{0.18\linewidth}
        \centering
        \includegraphics[trim={0cm 0cm 0cm 0cm}, clip, width=\linewidth]{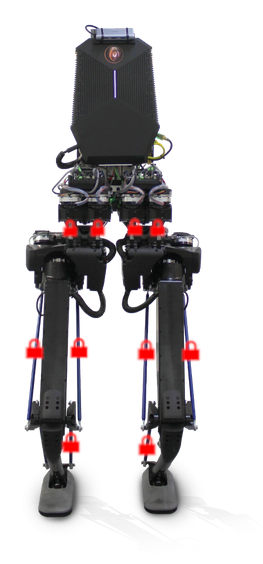}
    \end{subfigure}
    \hfill
    \begin{subfigure}{0.27\linewidth}
        \centering
        \includegraphics[width=\linewidth]{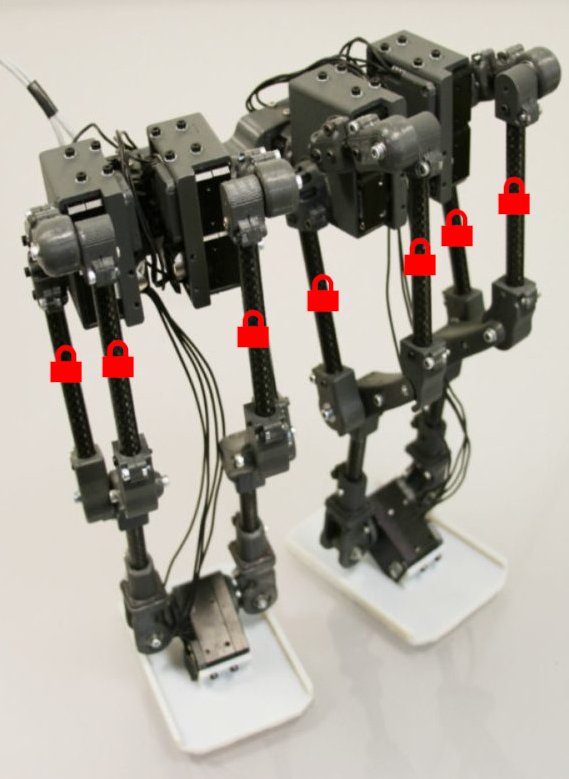}
    \end{subfigure}
    \hfill
    \begin{subfigure}{0.23\linewidth}
        \centering
        \includegraphics[trim={0cm 0cm 0cm 0cm}, clip, width=\linewidth]{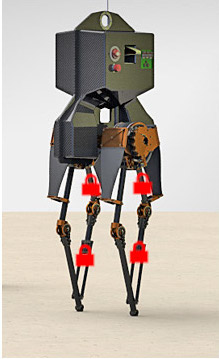}
    \end{subfigure}
    \hfill
    \caption{Examples of legged robots with closed-loop kinematics, ranging (top-left to bottom-right) from robots with a main serial chain (Fourier GR1 \cite{FourierGR1}, Unitree H1 \cite{HumanoidRobotG1_Humanoid}, Tesla Optimus \cite{TeslaOptimus}, Adam \cite{AdamPnd}), robot with an approximate serial chain (Digit \cite{AgilityProducts}) and robots without an approximate serial chain(Kangaroo \cite{roigHardwareDesignControl2022a}, Disney bipedal robot \cite{keving.gimDesignFabricationBipedal2018}, Atrias \cite{hubicki2016atrias}). Each red lock represents a visible closure of the kinematic chain.}
    \label{fig:teaser_robots}
\end{figure}
Several of the best modern biped robots already rely on such designs (see Fig~\ref{fig:teaser_robots}).
Yet, the specificity of this complex architecture is mostly ignored when generating their movements \cite{boukheddimi2023investigations}.

The approach mostly used in the literature is to only model an approximate serial chain in the generation of whole-body locomotion or movement, and to rely on an ad-hoc actuation model, often supported by the robot manufacturer, to transfer the reference joint motion and torques into actuator commands.
% \cite{s.kajitaBipedWalkingPattern2003,victorc.paredesResolvedMotionControl2022,hongkaidaiWholebodyMotionPlanning2014}.
On Atrias, a Spring Loaded Inverted Pendulum (SLIP) model is used in motion generation and computed torques are directly transferred to actuator control with of reduction ratio of 1 \cite{peekema2015template, ramezani2014performance}.
On Talos, Whole-Body controllers \cite{hongkaidaiWholebodyMotionPlanning2014}, such as Whole Body Model Predictive Control \cite{rawlings2017model} have been applied on an approximate serial model, letting the low level control transfer the serial torques to actuator commands \cite{ewendantecFirstOrderApproximation2022}.
Reinforcement Learning (RL) \cite{elliotchane-saneCaTConstraintsTerminations2024} methods, where the computation of the robot dynamic is delegated to a simulator, have also been applied to several of modern robots such as H1, Digit and Fourier GR1. 
Yet, modern GPU simulators used in RL, such as PhysicX or MuJoCo XLA (MJX), do not currently support closed kinematics.
This yield approaches where the complexity of the closed loop is only accounted for in the low level controller.
Only robot specific controllers, alternating the learning process and configuration projection, have yet been applied on such architecture, for instance \cite{guillermoa.castilloRobustFeedbackMotion2021} on Digit.\par
The main contribution of this paper is to derive a complete and general modeling methodology to enable the simulation of closed-loop kinematics, mostly targeting MPC although these approaches are the same as those used in the simulator for RL.
The dynamics of a poly-articulated robot is governed by the unconstrained equations of motion and founds efficient algorithms and their implementations in the literature \cite{royfeatherstoneRigidBodyDynamics2007, justincarpentierPinocchioLibraryFast2019, martinl.felisRBDLEfficientRigidbody2016}.
Recent works have proposed general methods to write the dynamic of a system under contact constraints \cite{justincarpentierProximalSparseResolution2021}, and showed a general form of its derivatives, \cite{carlosmastalliCrocoddylEfficientVersatile2019, shubhamsinghAnalyticalSecondOrderDerivatives2023}.
It is already accepted that closed-kinematic constraints can be cast under the same scope as contact constraints, to be used in the same framework \cite{justincarpentierProximalSparseResolution2021}.

In this paper, we first highlight the importance of accurately modeling closed-loop transmissions, as demonstrated in the experimental section.
To address this, our main contribution is to propose a comprehensive solution to compute the optimal movement for various walker kinematics. 
Our method applies efficiently to robots with an underlying serial kinematic chain (such as H1, electric-Atlas, and Adam), to those with approximated serial kinematics (such as Cassie and Digit), and extends to more complex kinematics without serial approximations (such as Kangaroo and the Disney biped). 
It relies on fully modeling the closed-loop linkages, including their derivatives, and integrating this model into a trajectory optimizer.
We experimentally show that our method generates optimal movements in case where neglecting actuator transmissions leads to sub-optimality, to unrealistic behavior, or even to motion generation failure. 
While simpler methods that decouple actuator models from kinematic models might be suitable for robots with approximated serial kinematics, we demonstrate that our method remains effective for more complex robots where such decoupling is not possible, opening the door to more efficient designs in the future.

After quickly recalling in Sec~\ref{sec:background} the background in optimizing the trajectory of robots subject to dynamic constraints, we provide an efficient computation of the derivatives of the dynamics of a robot involving closed-loop constraints in Sec.~\ref{sec:dynamic_derivatives}.
We then use this new dynamic to formulate and solve an optimal control problem for the walk of a robot with parallel actuation in Sec.~\ref{sec:experimental-study}).
The result section (Sec~\ref{sec:results}) highlight the limitations of using an \textit{Approximate Serial Model} of such a robot by comparing it to the \textit{Closed Kinematics Model} over different motions. 
We also demonstrate the application of our method to control a robot for which no \textit{Approximate Serial Model} can be defined.
\section{Background} \label{sec:background}
\subsection{Optimal control}
Optimal control of a multibody system consists in finding the control inputs that minimize a given cost function while satisfying the system dynamics. 
% In shooting approaches - single and multiple shooting - the problem is discretized in time and the control inputs are optimized over a finite time horizon.
In this paper, we seek for a solution to this problem by solving a trajectory optimization problem formulated by multiple shooting \cite{diehl:inria-00390435}, i.e. by optimizing over both the control inputs $u[k]$ and the states $x[k]$ at each discrete time step $k$, with the dynamics as an explicit constraint.
It is classically written as the following non-linear program (NLP):
\begin{equation} \label{equ:OCP_multiple_shooting_base}
    \begin{aligned}
        \min_{{u},\ {x}} & \sum_{k}^{N-1} l_k({x}[k], {u}[k]) + l_N({x}[N]) \\
        \text{s.t.} & \quad \forall k \in \llbracket 0, N-1\rrbracket \quad {x}{[k+1]} = f_k({x}[k], {u}[k]) \\
        % & \quad \forall k \in \llbracket 0, N-1\rrbracket \quad {c}_k({x}[k], {u}[k]) \leq {0} \\
        % & \quad {x}[0] = {x}_0
    \end{aligned}
\end{equation}
The cost function is usually defined as the sum of running costs $l_k$ and a terminal cost $l_N$.
The functions $f_k$ are defining the system dynamics at each time step $k$.
The case where $f$ models a poly-articulated system in contact is well-known and the derivations are recalled next.

\subsection{Multibody dynamics}
The dynamic of a multibody system is well described by the Lagrangian equation of motion:
\begin{equation}
    M(q)\ddot{q} + b(q, \dot{q}) = \tau(u) + J_c(q)^T \lambda
\end{equation}
where $M$ represents the generalized inertia matrix, $b$ the non-linear terms, $q$ the generalized coordinates of the system, $\dot{q}$ the generalized velocities\footnote{In our implementation, $q$ typically contains quaternions for representing basis orientation and ball-joints configurations, hence the notation $\dot{q}$, although classical, is abusive} and $\tau$ the torques applied on the system joints, function of the controls $u$ (usually, the controls correspond to the motor joints torques and all other joints torques are zero). 
The term $\lambda$ represents an external force applied on the system, where $J_c$ is the Jacobian of the contact point.
% In the context of closed kinematic chains, unknown forces are added to represent internal forces in the closure, that occurs at the contact points.
For systems subject to mechanical constraints (such as foot-ground contact or kinematics-closure), these forces $\lambda$ arise from the satisfaction of the constraints.

\subsection{Dynamics subject to constraints}
From Gauss principle of least action \cite{stephaneredonGaussLeastConstraints2002,davidbaraffFastContactForce1994,firdause.udwadiaEquationsMotionMechanical1996}, the acceleration ${\ddot{q}}$ of the system under contacts should be as close as possible to its free acceleration ${\ddot{q}_f}$, with respect the the kinematic metric, while satisfying the contact constraints.
This can be rewritten as the following optimization problem:
\begin{equation} \label{equ:contact_dynamic_optim_problem}
    \begin{aligned}
        \min_{\ddot{q}} & \quad \norm{\ddot{q} - {\ddot{q}_f}}_M^2 \\
        \text{s.t.} & \quad J_c \ddot{q} + a_0 = 0
    \end{aligned}
\end{equation}
The constraint in \eqref{equ:contact_dynamic_optim_problem} corresponds to the second-order time derivatives of the constraint of contact, where $a(q, \dot{q}, \ddot{q}) = J_c \ddot{q} + a_0(q, \dot{q})$ describes the relative acceleration of the contact points, that we consider here - without loss of generality - to be wanted to be zero, and where $a_0(q, \dot{q}) = a(q, \dot{q}, 0) = \dot{J_c}\dot{q}$ is the acceleration due to the velocity only. 
Deriving first-order optimality conditions, \eqref{equ:contact_dynamic_optim_problem} boils down to:
\begin{equation} \label{equ:contact_dynamic_KKT_system}
    \underbrace{\begin{bmatrix}
        0 & J_c \\
        J_c^T & M
    \end{bmatrix}}_K
    \underbrace{\begin{bmatrix}
        \lambda \\
        \ddot{q}
    \end{bmatrix}}_y
    =
    \underbrace{\begin{bmatrix}
        -a_0 \\
        \tau(u) - b 
    \end{bmatrix}}_k
\end{equation}
where $f$, the dual variables of the optimization problem, correspond to the contact forces applied on the system.
This system can be solved to get $\ddot{q}$ and $\lambda$ given the state $x=\begin{pmatrix}
q& \dot{q}
\end{pmatrix}$ and the controls $u$.
The solution $y = K^{-1}k$ to this problem exists and is unique whenever the matrix $K$ is invertible, which usually happens when the matrix $M$ is positive definite and when the constraints are not redundant.
Proximal resolution has been proposed to solve the problem in settings where $J_c$ is not full rank \cite{justincarpentierProximalSparseResolution2021}.
This formulation has been extensively applied to model ground contact of walking robots. In the next section, we propose to modify the constraint \ref{equ:contact_dynamic_optim_problem} to model internal forces arising in closed-kinematics linkages. 

\subsection{Derivatives of the constrained dynamics}
MPC typically uses gradient-based solvers to get the solution of \eqref{equ:OCP_multiple_shooting_base}, which implies to evaluate the derivatives of $y$ with respect to $x$ and $u$.
In \cite{carlosmastalliCrocoddylEfficientVersatile2019} the authors proposed the derivatives of a robot in contact with its environment. 
More recent work generalize these to arbitrary contacts \cite{shubhamsinghAnalyticalSecondOrderDerivatives2023, justincarpentierPinocchioLibraryFast2019}.
Following these works, the gradient of $y$ with respect to $z \in \{q, \dot{q}, u\}$ can be reduced to:
\begin{equation} \label{equ:gradient_y}
    \frac{\partial y}{\partial z} = K^{-1} \begin{bmatrix} 
        \frac{\partial\ {a_0}}{\partial {z}} \\
        \frac{\partial\ {ID}}{\partial {z}}
    \end{bmatrix}
\end{equation}
where the Inverse Dynamics (ID) function outputs the joint torques creating acceleration $\ddot{q}$ under contact forces $f$.
\begin{equation}
    {ID}(q, \dot{q}, \ddot{q}, \lambda) = M {\ddot{q}} + J_c^T \lambda + b
\end{equation}

The derivatives of these terms with respect to the controls $u$ will not be explicited in this paper as the first term is independent on $u$ and the second depends on the chosen actuation model.
The derivatives of ID when $J_c$ are the joint jacobians have been established \cite{justincarpentierAnalyticalDerivativesRigid2018} and are sufficient for the case of a contact between the robot and its environment, yet we will see that they need to be extended in our setting. \\

\section{A constraint modeling the kinematic closure} \label{sec:dynamic_derivatives}

\subsection{Formulation of the constraint in acceleration}
We consider the mechanical linkage between two bodies of the robot, characterized by frames $\F_1$ and $\F_2$ rigidly attached to each of them.
We choose to write the 6D contact constraint on the relative placement as:
\begin{equation}
    \label{equ:contact_constraint}
    \begin{aligned}
        \Log{({}^{1}M_{2})} &= 0
    \end{aligned}
\end{equation}
where ${}^{1}M_{2} \in {\SE3}$ is the rigid transformation between the frames and $\Log$ is the retractation from $\SE3$ to $\real{6}$ \cite{joansolaMicroLieTheory2018}.\\
% In the following, we will note $1$ and $2$ instead of $c_1$ and $c_2$ for ease of notation.
The first-order time derivative of this constraint can be expressed as the difference in spatial velocities expressed in a common frame.
We choose as a convention to express all quantities in the frame $\F_1$, giving the constraint:
\begin{equation}
    \label{equ:contact_velocity}
        \nu_c = {}^1\nu_1 - ^1X_2 {}^2\nu_2 = 0
\end{equation}
where, following the notations introduced by Featherstone \cite{royfeatherstoneRigidBodyDynamics2007}, ${}^1\nu_1$ and ${}^2\nu_2$ are the spatial velocities of the two bodies expressed in their respective frames, ${}^1X_2$ is the Plücker coordinate transform, i.e. the adjoint matrix of $\SE3$, from $\F_2$ coordinates to $\F_1$ coordinates.
In the same way, the second-order time derivative of \eqref{equ:contact_constraint} can be expressed as the derivative of the relative spatial velocity $\nu_c$.
\begin{equation} \label{equ:contact_acceleration}
    \begin{aligned}
        a_c &= {}^1a_1 - ^1X_2 {}^2a_2 - \cross{^1\nu_2 - ^1\nu_1}{}^1\nu_2\\
        &= \underbrace{{}^1a_2}_{\gamma_1} - \underbrace{{}^1X_2 {}^2a_2}_{\gamma_2} + \underbrace{\cross{^1\nu_1}^1\nu_2}_{\gamma_3} 
    \end{aligned}
\end{equation}
with the spatial cross-product $\cross{\nu}$ (\textit{small adjoint}) given by
\begin{equation}
    \cross{\nu} \triangleq \begin{bmatrix}
        \cross{\bf{\omega}} & \cross{\bf{v}} \\
        \bf{0} & \cross{\bf{\omega}}
    \end{bmatrix}
\end{equation}
% and all motion quantities are written with a "linear then angular" convention.\\

\subsection{Differentiation of the acceleration constraint}
As shown in \eqref{equ:gradient_y}, we need the derivatives of the terms $\gamma_1$, $\gamma_2$ and $\gamma_3$ with respect to the configuration vector $q$ and velocity $\dot{q}$.
The first of these derivatives yields directly:
\begin{equation}
    \label{equ:dgamma1_dq}
    \begin{aligned}
        \frac{\partial \gamma_1}{\partial q} &= \frac{\partial {}^1a_1}{\partial q} =  \frac{{}^1 \partial a_1}{\partial q}\\
    \end{aligned}
\end{equation}
Note that we make here a clear distinction between the terms $\frac{\partial {}^Aa_B}{\partial q}$ and $\frac{{}^A \partial a_B}{\partial q}$ as explained in \cite{royfeatherstoneRigidBodyDynamics2007} (section 2.10).
% , even if those are equal in this case.
To compute the derivative of the second term, we can use the method proposed in \cite{sebastienkleffDerivationContactDynamics} - i.e. we search the time derivatives of $\gamma_2$ under the form $\dot{\gamma}_2 = G \dot{q}$ to deduce $\frac{\partial \gamma_2}{\partial q} = G$.
\begin{equation}
    \label{equ:dgamma2_dt}
    \begin{aligned}
        \frac{\partial \gamma_2}{\partial t} &= \frac{\partial {}^1X_2.^2a_2}{\partial t}\\
        &= \frac{\partial {}^1X_2}{\partial t}{}^2a_2 + {}^1X_2\frac{\partial {}^2a_2}{\partial t}\\
        &= \cross{{}^1(\nu_2 - \nu_1)}\phantom{.}^1X_2{}^2a_2 + {}^1X_2\frac{{}^2 \partial a_2}{\partial t}\\
        &= -\cross{{}^1X_2.^2a_2}{}^1(\nu_2 - \nu_1) + {}^1X_2\frac{{}^2 \partial a_2}{\partial t}\\
    \end{aligned}
\end{equation}
which yields
\begin{equation}
     \frac{\partial \gamma_2}{\partial q} = -\cross{{}^1X_2{}^2a_2}({}^1J_2 - {}^1J_1) + {}^1X_2\frac{{}^2 \partial a_2}{\partial q}
\end{equation}
We proceed in a similar way for the third and last term:
\begin{equation}
    \label{equ:dgamma3_dt}
    \begin{aligned}
        \frac{d \gamma_3}{d t} &= \frac{\partial \cross{^1\nu_1}{}^1X_2{}^2\nu_2}{\partial t}\\
        &= \cross{\frac{\partial ^1\nu_1}{\partial t}}^1\nu_2 + \cross{^1\nu_1}(\frac{\partial {}^1X_2}{\partial t}{}^2\nu_2) + \cross{^1\nu_1}({}^1X_2\frac{\partial {}^2\nu_2}{\partial t})\\
        % &= -\cross{^1\nu_2} \frac{{}^1 \partial \nu_1}{\partial t} + \cross{{}^1\nu_1}(\cross{{}^1(\nu_2 - \nu_1)}\phantom{.}^1X_2{}^2\nu_2) \\ &\hspace{3em} + \cross{^1\nu_1}{}^1X_2\frac{\partial \phantom{.}^2\nu_2}{\partial t}\\
        % &= -\cross{^1\nu_2} \frac{{}^1 \partial \nu_1}{\partial t} - \cross{{}^1\nu_1}\left(\cross{{}^1X_2{}^2\nu_2}({}^1\nu_2 - {}^1\nu_1)\right) \\ &\hspace{3em} + \cross{^1\nu_1}\phantom{.}^1X_2\frac{{}^2\partial \nu_2}{\partial t}\\
    \end{aligned}
\end{equation}
which gives after development
\begin{equation}
    \label{equ:dgamma3_dq}
    \begin{aligned}
        \frac{\partial \gamma_3}{\partial q} &= -\cross{^1\nu_2} \frac{\phantom{.}^1 \partial \nu_1}{\partial q} - \cross{{}^1\nu_1}\left(\cross{{}^1X_2.^2\nu_2}({}^1J_2 - {}^1J_1)\right) \\ &\hspace{3em} + \cross{^1\nu_1}\phantom{.}^1X_2.\frac{\phantom{.}^2\partial \nu_2}{\partial q}
    \end{aligned}
\end{equation}
All these terms can accessed using rigid body dynamics algorithms such as those included in Pinocchio \cite{justincarpentierPinocchioLibraryFast2019}.\\
As the action matrices do not depend on $\dot{q}$, the derivatives of $a_c$ with respect to $\dot{q}$ are direct and left to the reader.

\subsection{Derivatives of Inverse Dynamic}
We will now look at the derivatives of ID with respect to $q$ and $\dot{q}$. 
% To solve for these, we can observe that the expression of $\tau$ resembles the output of the Recursive Newton-Euler Algorithm (RNEA) for which derivatives have already been computed \cite{justincarpentierAnalyticalDerivativesRigid2018}.\\
% However, this algorithm assumes that the contact forces applied on the joints are independent of the configuration vector $q$ and on the articular velocity $\dot{q}$. While the latter remains correct for closed loop contact constraints, the former is not as the contact force changes direction based on the robot configuration.
% To link the expected derivatives to the ones of the RNEA, let us write first the expression as a function of the forces applied on the joints.
Let us write the expression of ID as a function of the forces applied on the joints.
We will denote these forces by $\phi_k$ and the corresponding joint Jacobians $J_k$ for joint $j_k$.
\begin{equation}
    \label{equ:RNEA_expression_detailled}
    \begin{aligned}
        ID(q, \dot{q}, \ddot{q}, f) &= M({q}){\ddot{q}} + {b}({q}, {\dot{q}}) + \sum_k J_k(q)^T \phi_k(q, f) \\
    \end{aligned}
\end{equation}
where the $\phi_k$ are typically expressed in the reference frame of joint $j_k$ denoted by $\F_{j_k}$.
We will now omit the dependences to simplify the notation.
The derivative with respect to the $q$ is given by:
\begin{equation}
    \label{equ:RNEA_derivative_q}
    \begin{aligned}
        \frac{\partial {ID}}{\partial q} &= \frac{\partial (M\ddot{q} + b)}{\partial q} + \sum_k \frac{\partial J_k^T}{\partial q}\phi_k + \sum_k J_k^T\frac{\partial \phi_k}{\partial q} \\
    \end{aligned}
\end{equation}
The first two terms are the classical Recursive Newton-Euler Algorithm (RNEA) derivatives, which are well established and implemented in the Pinocchio library \cite{justincarpentierAnalyticalDerivativesRigid2018,justincarpentierPinocchioLibraryFast2019}.\par
To compute the third term, we denote by $j_1$ and $j_2$ the parent joints of $\F_1$ and $\F_2$.
Following the choice of \eqref{equ:contact_acceleration}, the force $f$ arising from the constraint \eqref{equ:contact_constraint} is expressed in $\F_1$.
Considering that only $f$ acts on the system, then all forces $\phi_k$ are null except $\phi_1$ and $\phi_2$ which are:

\begin{equation}
    \label{equ:contact_force_on_joints}
    \begin{aligned}
        \phi_1 &= \phantom{.}^{j_1}X_{c_1}^*f\\
        \phi_2 &= 
        % -\phantom{.}^{j_2}X_{c1}^*(q)f\\
            -{}^{j_2}X_{c_2}^*{}^{c_2}X_{c_1}^*(q)f
    \end{aligned}
\end{equation}
where $X^*$ is the Plücker transform on forces (dual adjoint), ${}^{j_1}X_{c_1}$ is the fixed placement of the contact frame $\F_1$ with respect to the joint frame $\F_{j_1}$ (respectively ${}^{j_2}X_{c_2}$) and ${}^{c_2}X_{c_1}$ is function of $q$.
We can see that $\phi_1$ is independent of $q$ while $\phi_2$ is not.
Its derivative with respect to $q$ is given by:
\begin{equation}
    \label{equ:phi2_derivative}
    \begin{aligned}
        \frac{\partial \phi_{j_2}}{\partial q} &= - \frac{\partial \phantom{.}^{j_2}X_{c_1}^*}{\partial q}f\\
        &= \cross{{}^{j_2}X_{c_1}^*f}^*(J_2 - {}^{j_2}X_{j_1}J_1)
    \end{aligned}
\end{equation}
where, $J_1$, $J_2$ are the joint jacobians respectivelly expressed in $\F_1$ and $\F_2$ and the term $\cross{f}^*$ can be defined as follows:
\begin{equation}
    \label{equ:cross_force}
    \begin{aligned}
        \cross{f}^* &\triangleq \begin{bmatrix}
            0 & \cross{f_{linear}}\\
            \cross{f_{linear}} & \cross{f_{angular}}
        \end{bmatrix}
    \end{aligned}
\end{equation}
With the existing derivatives of RNEA \cite{justincarpentierAnalyticalDerivativesRigid2018}, this completes the computation of the derivative of ID with respect to $q$.

\section{Benchmark Implementation} \label{sec:experimental-study}
\subsection{Kinematic models}
We consider two different models of robots with closed-loop kinematics.\\
\subsubsection{Robot with an approximate serial kinematics}
First, we consider a robot that is composed of a main serial kinematic chain, whose actuators are shifted upfront of their serial joints by closed-loop mechanical linkages, similar to H1, Adam or electric Atlas.
For these robots, a common approach \cite{Mronga2022WBC} is to decouple the generation of the motion of the main serial chain from the control of the shifted actuators.
We will show that not accounting for the actuator model in the motion generation leads to serious failure cases when the robot is pushed beyond slow walk on a flat terrain.
\begin{figure}
    \centering
    \hfill
    \begin{subfigure}{0.28\linewidth}
        \centering
        \includegraphics[trim={3cm 1.7cm 1cm 1cm}, clip, width=1.2\linewidth]{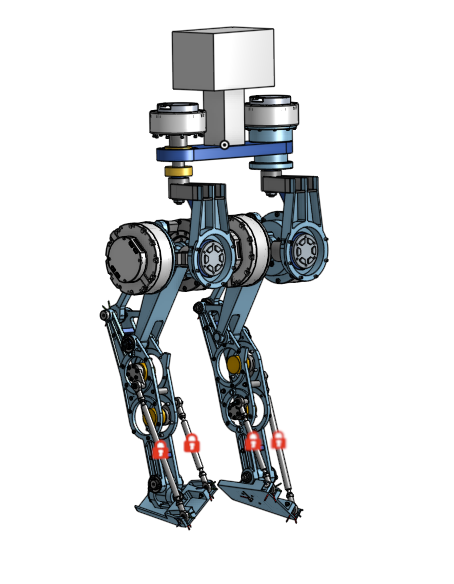}
    \end{subfigure}
    \hfill
    \begin{subfigure}{0.28\linewidth}
        \centering
        \includegraphics[trim={1cm 0cm 1cm 0cm}, clip, width=\linewidth]{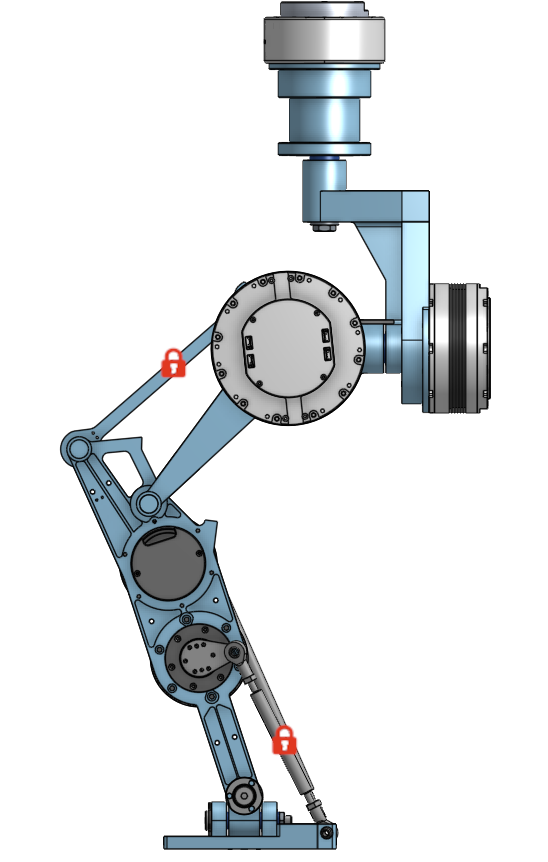}
    \end{subfigure}
    \hfill
    \caption{Robot model used for our benchmark. Each red lock represents a closure of the kinematic chain. 
    In our model, we represent the chain as a tree-like structure with added contact constraints by splitting the bar in two at the lock position and adding 6D contact constraints}
    \label{fig:robot}
\end{figure}
For this first benchmark, we used the model of the robot Bipetto, available in open source \cite{GithubParallelRobots} with explicit actuator model and depicted in Fig \ref{fig:robot}.
Like H1, Adam, or electric Atlas, it features a knee motor with a four-bar linkage and two ankles motors with intricate four-bar linkages, inspired by Digit.
Each closed-loop transmission creates a reduction ratio between the motors and the joints that depends on the robot configuration.
\begin{figure}
    \centering
    \includegraphics[width=0.65\linewidth]{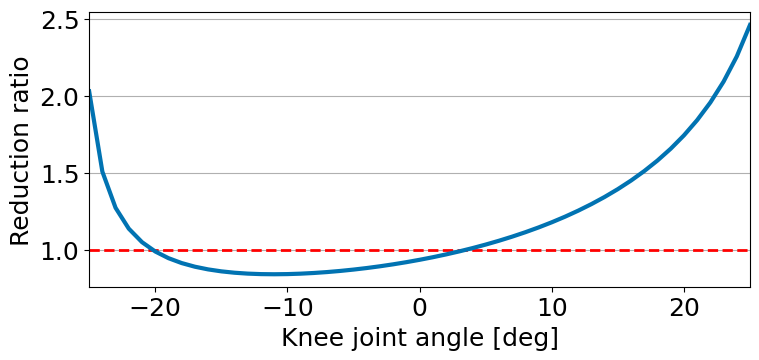}
    \caption{Variation of the reduction ratio of the knee actuation with respect to the knee angle.
    The 0 angle corresponds to a nominal configuration of the robot while positive angles correspond to a stretched leg.}
    \label{fig:reduction_ratio}
\end{figure}
We show in Fig. \ref{fig:reduction_ratio} the variation of the reduction ratio of the knee actuation with respect to the knee angle, revealing an highly non-linear relation.
Getting an analytical formula for this variable reduction ratio is in general very difficult and poorly generalizes to new models, limiting its possible use in control settings. 
Our method overcomes this limitation by providing a general formulation, directly transferable to other robot models.

Our benchmark compares the \textit{Closed-Kinematics} model including closed-loop linkages against an \textit{Approximate Serial} model considering only the serial joints of the robot (hip, knee, and ankle joints) and freezing the other degrees of freedom corresponding to the motor-to-joint transmission.
In this simplification, the non-serial motors are fixed and fictive actuation is added on serial joints, yielding a fully serial model with 6 directly actuated revolute joints per leg.
The inertia of the closed-loop transmission is modeled to be as realistic as possible, making no assumptions of low inertia. In the \textit{Approximate Serial} model, the linkages joints are blocked in the initial configuration, yielding minimal inertia variations compared to the \textit{Closed-Kinematics} model.\\
We first exhibit the limitation of not accounting for the actuation in the OCP when a \textit{Approximate Serial} model can be defined.
We compare the movements generated with our OCP including the full \textit{Closed-Kinematics} model against a classical OCP \cite{ewendantecFirstOrderApproximation2022} using the \textit{Approximate Serial} model and neglecting the actuation.
In both OCP, the costs are the same, as described in Sec. \ref{subsec:OCP_costs}.
To compare the two movements, we simply lift the movement of the \textit{Approximate Serial} model to the full \textit{Closed-Kinematics} model, using a procedure described in Appendix \ref{app:lifting}.
Of course, this procedure may fail if the \textit{Approximate Serial} trajectory corresponds to infeasible configuration or torques.
% We then show that our method extends to more complex kinematics using the same OCPs with a second robot, for which no \textit{Approximate Serial} model exists.

\subsubsection{Robot without underlying serial kinematics}
We show that our method extends to robots without a simple correspondence to a serial chain, such as the Disney biped \cite{keving.gimDesignFabricationBipedal2018} or Kangaroo \cite{roigHardwareDesignControl2022a}.
Note that Digit \cite{AgilityProducts} also formally enter in this category due to its shin joint which is often mistaken as a knee joint, leading to more significant approximation than for example H1 \cite{guillermoa.castilloRobustFeedbackMotion2021}.
\begin{figure}
    \center
    \includegraphics[width=0.45\linewidth]{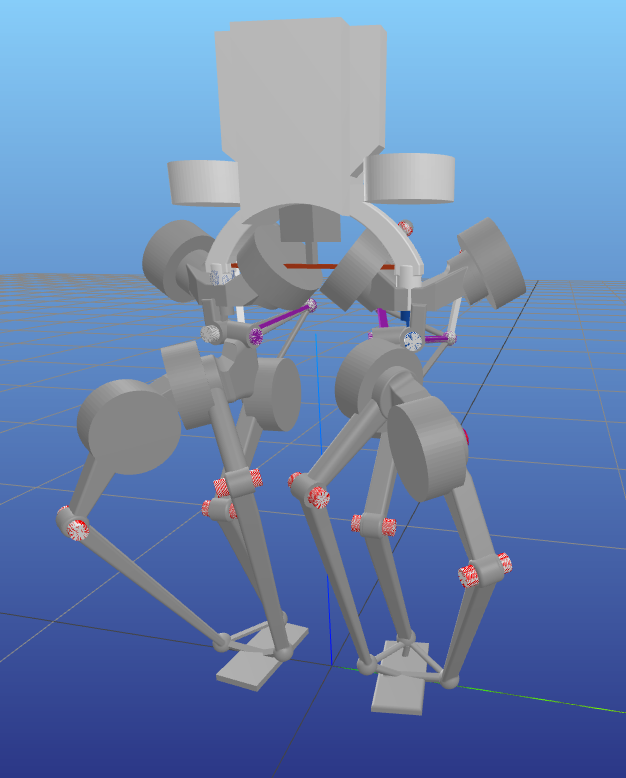}
    \caption{Model of the Cleobot \cite{batto:hal-04717159}, a fully parallel robot. It uses three parallel chains of equivalent inertia for the knee and ankle actuation and three other parallel chains for hip actuation}
    \label{fig:cleobot}
\end{figure}
We used the model of the Cleobot walker \cite{batto:hal-04717159}, presented in Fig~\ref{fig:cleobot}, also available in open source, whose lower leg is composed of 3 parallel bodies attached to 3 revolute joints in place of the knee and ankle, and for which a serial model would be irrelevant to approximate.

\subsection{Optimal control problem details and implementation} \label{subsec:OCP_costs}
We implement several variation of a base OCP pattern to achieve squats, walk, stairs and jump tasks.
The dynamics under closed-loop constraints and its derivatives have been implemented in C++ in the Crocoddyl library \cite{carlosmastalliCrocoddylEfficientVersatile2019}, used to write and solve the OCPs \cite{GithubCrocoddylFork}.

\paragraph{Contact patterns}
In our OCP structure, we must define the contact pattern, i.e. when each foot is in contact with the floor, in order to associate to each time step the correct dynamics.
For the squat task, foot are always in contact. The OCP for the walk motion consists of 4 steps of alternating double support and single support phases \cite{ewendantecFirstOrderApproximation2022}. For the jump OCP, we define a simple sequence composed of double supports and a flying phase of known duration.

\paragraph{Regularization costs}
For all tasks, we regularize the controls and the states around zero and around a reference configuration respectively.
The ground contact forces are also regularized around 0 when the foot is in the air, the weight of the robot when the foot is in contact with the ground in single support phases and half of it during double support phases.

\paragraph{Impact costs}
As the dynamics of the robot under contact only constraint the relative accelerations of the contact points, we add soft constraints on the velocity and placement of the foot at the end of single support phases to ensure it lands flat, still, and at the correct height. 
Not that hard constraints could be used instead using more recent solvers \cite{wilsonjalletPROXDDPProximalConstrained, armandjordanaStagewiseImplementationsSequential} in order to limit the amount of tuning.

\paragraph{Target costs}
The target costs are more specific to each task and, with the contact pattern, define in the OCP the task to be performed.
First, the walk motion is defined by a target center of mass (CoM) velocity.
It is represented in the OCP as a running cost with a residual corresponding to the CoM velocity error and as a terminal cost with a residual corresponding to the difference, in the forward direction, between the expected displacement given the target velocity and the real terminal position.\\
The squats motions are defined as a vertical translation of the CoM.
Given a target minimal CoM elevation, we define a sinusoidal trajectory for the CoM elevation, starting with 0 velocity, reaching the target then returning to its initial position and ending with 0 velocity. 
In the OCP, this results in a running cost on the CoM elevation (planar CoM motions are not penalized), with a residual corresponding to the deviation from the target trajectory.\\
For the jump task, we proceed by defining a flying duration and deducing the expected maximum elevation of the CoM, that should be attained at mid flight.
We then only apply in the OCP a cost for the CoM elevation at this time step, with a residual defined as the difference between the expected elevation and the actual one.
In opposition to the squat task, this cost at only one time step is sufficient to constrain the entire flight phase.\\
Finally, the stairs climbing motion is defined by both a CoM velocity and a slope (i.e. a sequence of steps height).
It simply adds to the walk OCP, a cost at each impact time (i.e. at the switch between single support to double support), with a residual defined as the difference between the foot elevation and the target step height.
\paragraph{Additional costs}
Other costs are added to improve the realism of the trajectory (foot fly-high, center of pressure, forces in friction cones...). We do not present them here but one can find their definition in the example codes.
Lastly, in the walking experiment, a cost penalizes the drift of the COM from its initial height.
% The weight of this cost will be changed to emphasize differences of behaviors between models by allowing the use of the non-linear part of the reduction ratio.

\paragraph{Actuator limits}
Our method can directly handle the real actuator constraints (which would not be possible when decoupling the whole-body motion generation from the actuators control).
In the following section, we present the results without actuator limits, to better emphasize the comparison and prevent conservative assumptions.

\section{Results} \label{sec:results}
\subsection{Comparison of \textit{Approximate Serial} and \textit{Closed-Kinematics} models}
\subsubsection{Squats}
\label{subsec:squats}
\begin{figure}
    \center
    \includegraphics[width=\linewidth]{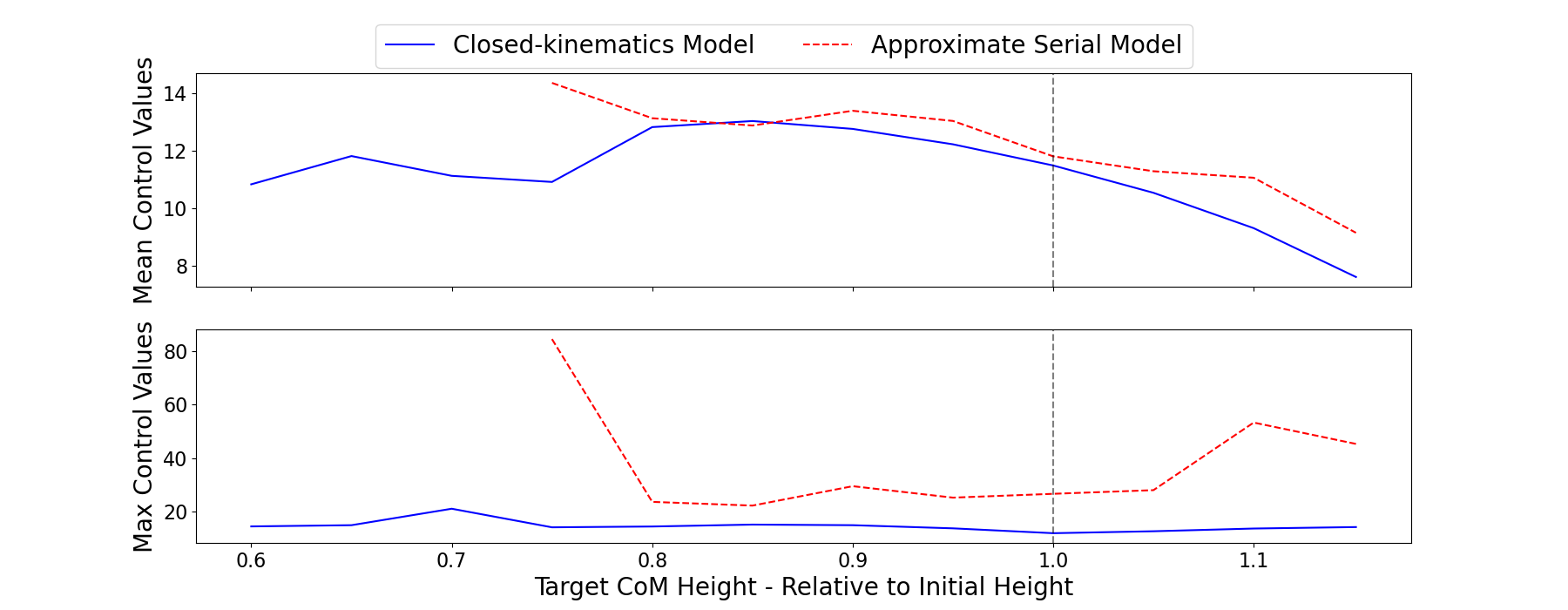}
    \caption{Evolution of the Maximum and Mean knees controls as functions of the target CoM elevation in the squat motion. The target elevation is noted relative to the initial CoM height. For deeper squats, the required joint torques for the \textit{Approximate Serial} trajectory yield excessive motor torques}
    \label{fig:controls_vs_com_elevation}
\end{figure}

Both problems converge to a reasonable squat movement for different target elevations, varying from 0.5 to 1.2 times the initial CoM elevation.
Yet, without a proper model of the closed-loop transmission, the \textit{Approximate Serial} trajectory reaches a singular configuration where the reference joint torques cannot be produced by the motors, as shown in Fig~\ref{fig:controls_vs_com_elevation}.
On the opposite, the \textit{Closed Kinematics} trajectory anticipates and produces an effective movement.
While this effect is expected for such simple scenario, we will now show that it similarly appears in more complex movements.

\subsubsection{Flat ground walk}
\begin{figure}
    \center
    \includegraphics[width=\linewidth]{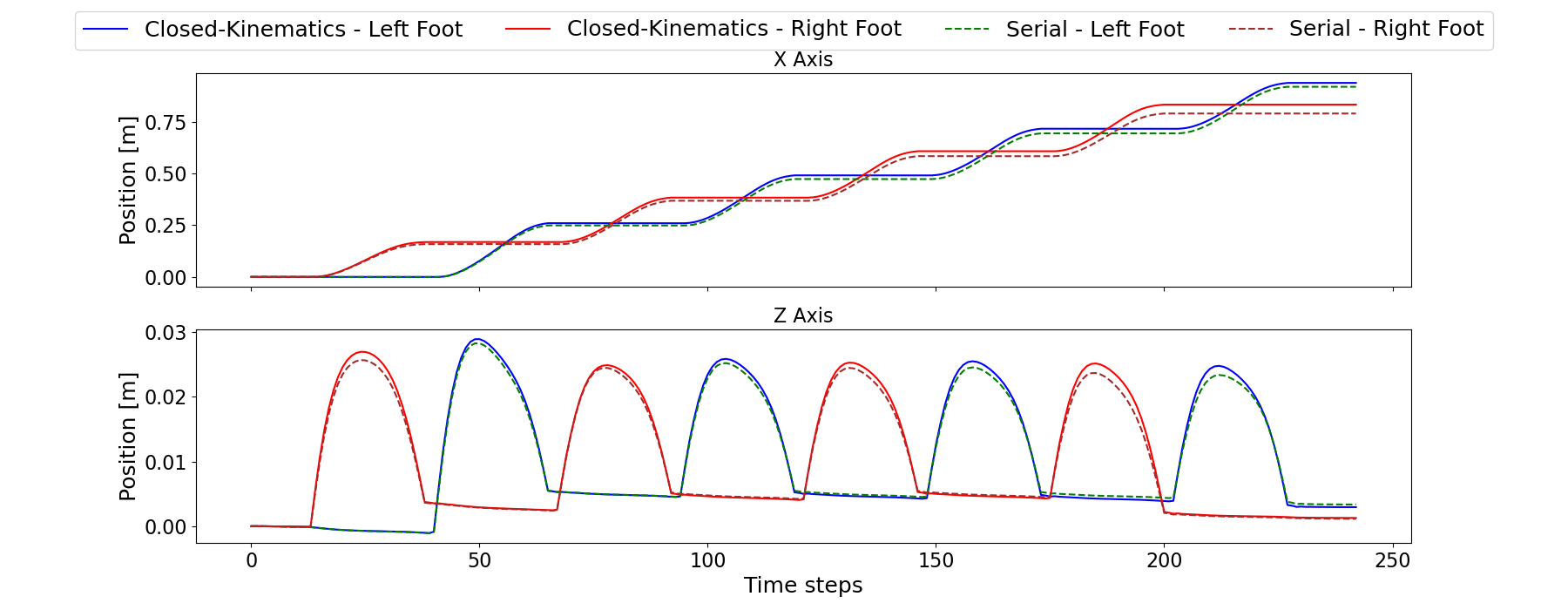}
    \caption{Feet trajectory for the \textit{Approximate Serial} and \textit{Closed-Kinematics} trajectories during the reference walk motion}
    \label{fig:flat_walk_foot}
\end{figure}

\begin{figure}[t]
    \center
    \includegraphics[width=\linewidth]{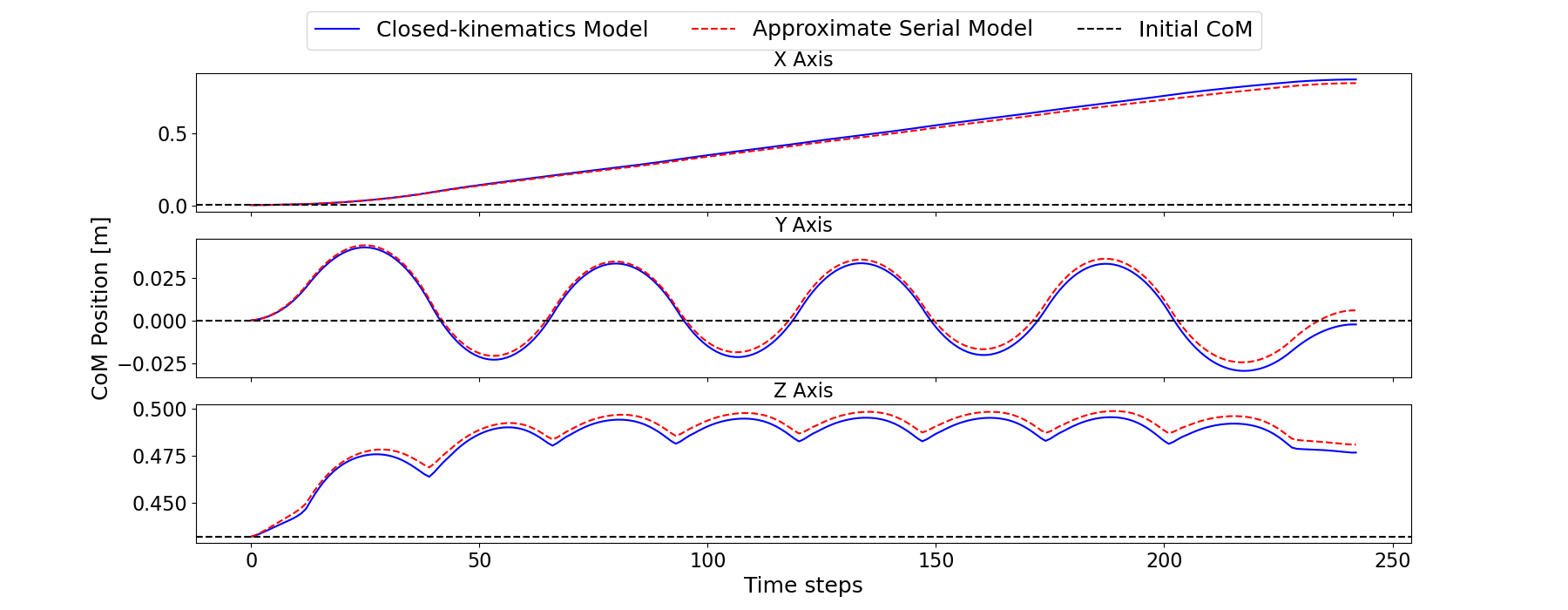}
    \caption{CoM position trajectory for the \textit{Approximate Serial} and \textit{Closed-Kinematics} trajectories during the reference walk motion. The CoM reaches different steady state positions, revealing differences in the optimal trajectories}
    \label{fig:flat_walk_com_pos}
\end{figure}

We now consider a walking motion on a flat terrain with a constant speed at $0.5 m/s$ (i.e. equivalent to 6km/h walk for a human-sized robot), with an initial robot configuration that places its base at $0.575\ m$ above the ground.
Fig~\ref{fig:flat_walk_foot} emphasizes that the movements obtained with both models are quite similar in appearance.
Yet, the trajectory of the CoM, presented in Fig~\ref{fig:flat_walk_com_pos}, show small deviations, especially in the elevation (i.e. position along the Z-axis).
This reveals a convergence toward a slightly different optimal trajectory, yet not yielding any detrimental behavior for the \textit{Approximate Serial} model.

\subsubsection{Variation of the velocity command}\label{subsec:vel_variation}
Changing the velocity command from $0.0 m/s$ to $1.2 m/s$ (keeping the contact pattern unchanged) tends to accentuate the previous behavior.
\begin{figure}
    \center
    \includegraphics[width=\linewidth]{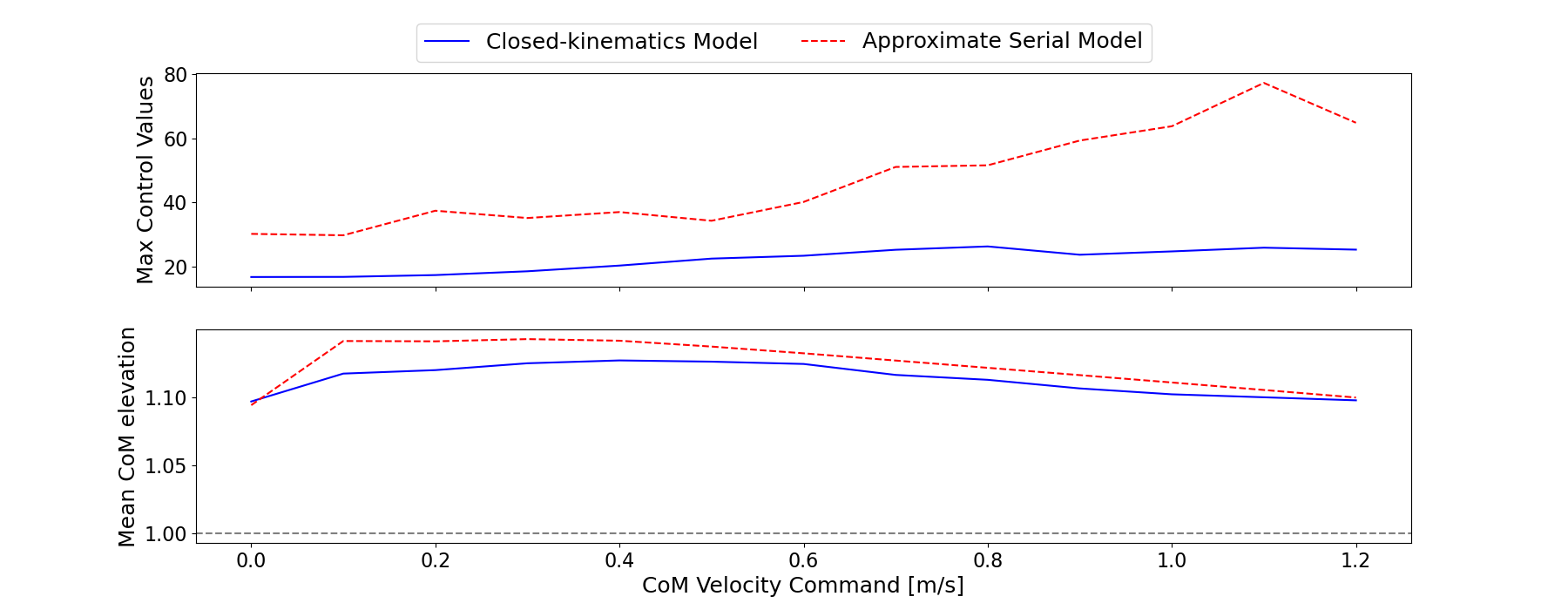}
    \caption{Elevation of the CoM of the robot and of the mean knees motors controls as functions of the target forward velocity. The trajectory of the Approximate Serial Model yields excessive motor torques, eventually leading to convergence failure in the lifting process \ref{equ:open_to_closed_problem}}
    \label{fig:velocity_compos}
\end{figure}
Figure \ref{fig:velocity_compos} shows the maximum knee controls reached over the steady state portion of the walking motion along with the mean CoM elevation.
While we observe that the CoM elevation only diverge slightly for both models, a large difference can be observed in the peak controls.
The contact sequence for the model being fixed, an higher velocity must be reached by using wider steps.
When walking with small steps, the \textit{Approximate Serial} model tend to lead to an higher CoM elevation, allowing it to reduce the joint torques.
Yet, accelerating the walk forces the robot to have stretched legs during the steps, lowering the CoM, and eventually converging to a similar CoM elevation for both models.
This pushes the closed-loop transmission in its non-linear part and the \textit{Approximate Serial} model then fails to account correctly for the reduction ratio, leading to exploding motor control and yielding a trajectory that cannot be transferred to the complete model (this occurs at $0.9\ m/s$ and above for our contact pattern).
This demonstrates that the limits of the \textit{Approximate Serial} model observed on the squat motion (Sec \ref{subsec:squats}) also appear on some more complex motions.
More generally, the \textit{Closed-Kinematics} model is able to take advantage of the actuator variable reduction created by the parallel linkage, hence leading to reasonable motor effort independently of the walk speed.

\subsubsection{Climbing stairs and jumping}
We generalized the results to more general locomotion scenarios: climbing stairs and jumping, with various robots. Like for fast walking, the \textit{Approximate Serial} model fails to properly anticipate the limits of the actuation linkages and results in trajectories with excessive torques or reaching singularities.
When jumping, mostly the knee reaches over extension and could be artificially clamped or penalized.
Yet our method makes that hyper-parameter tuning useless, providing a better generalization. 
For stair climbing, clamping the knee would be even more limiting and could completely make some motions unfeasible.
Moreover, for stair climbing, the ankle limits are also reached, for which an ad-hoc limitation (with 2 degrees of freedom) is more difficult to decide. 
The movements are reported in the companion video for the Bipetto walker and Digit.

\subsection{Robot without underlying serial kinematics}
\begin{figure}[t]
    \centering
    \hfill
    \begin{subfigure}{0.45\linewidth}
        \centering
        \includegraphics[trim={18cm 8cm 18cm 3cm}, clip, width=\linewidth]{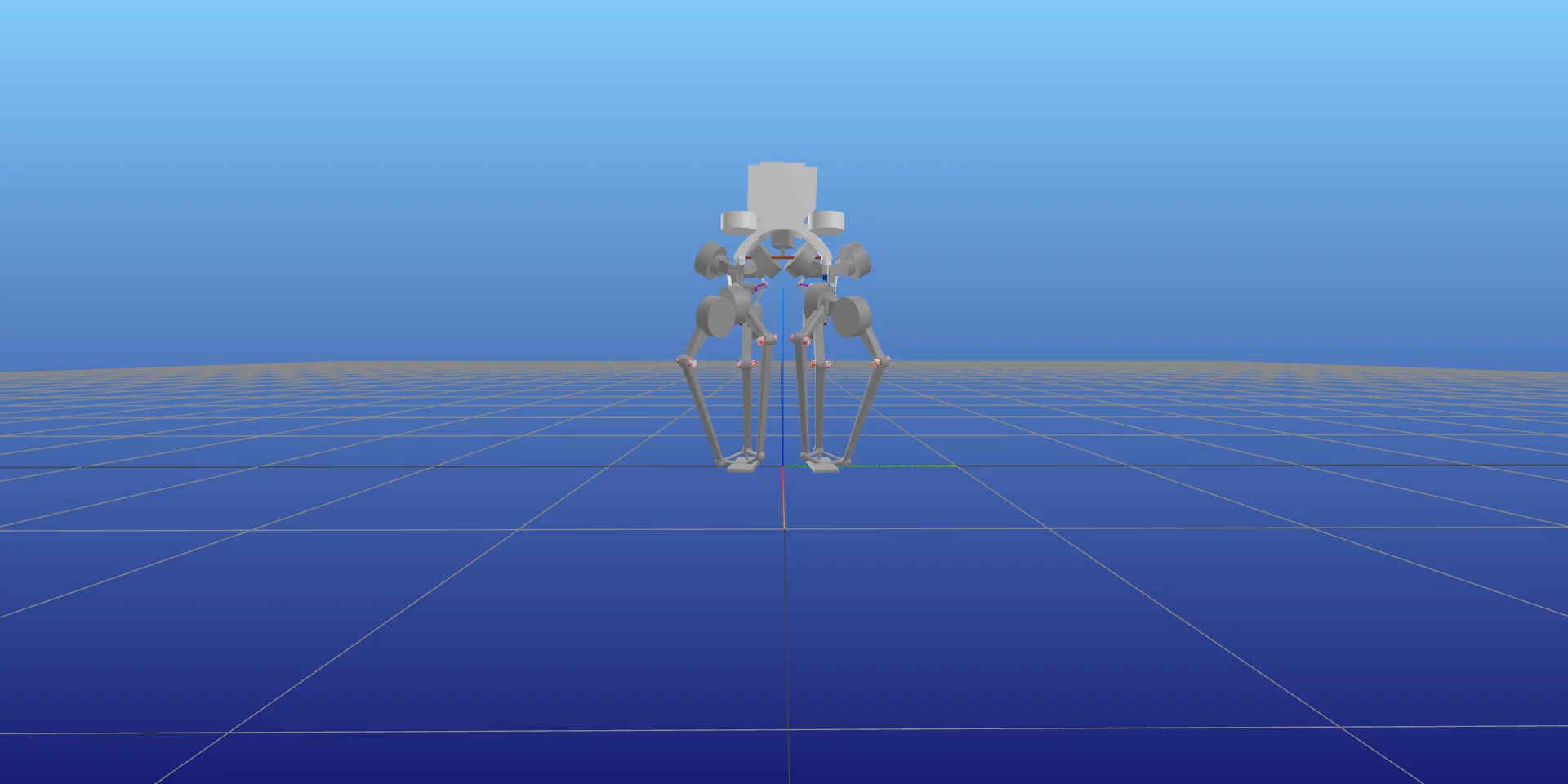}
        \caption{Standing}
    \end{subfigure}
    \hfill
    \begin{subfigure}{0.45\linewidth}
        \centering
        \includegraphics[trim={18cm 8cm 18cm 3cm}, clip, width=\linewidth]{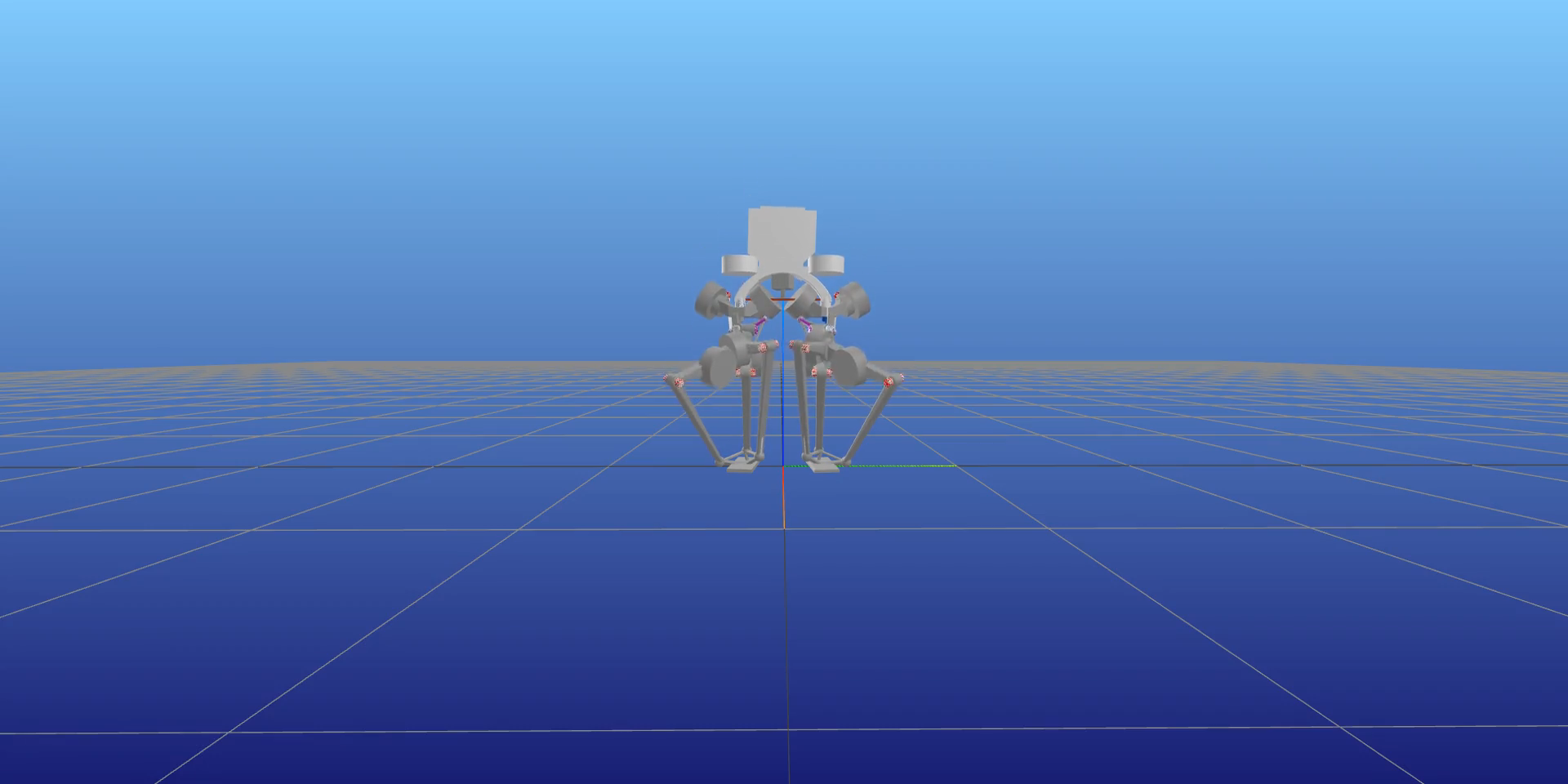}
        \caption{Squatting}
    \end{subfigure}
    \hfill
    \\
    \hfill
    \begin{subfigure}{0.45\linewidth}
        \centering
        \includegraphics[trim={15cm 8cm 21cm 3cm}, clip, width=\linewidth]{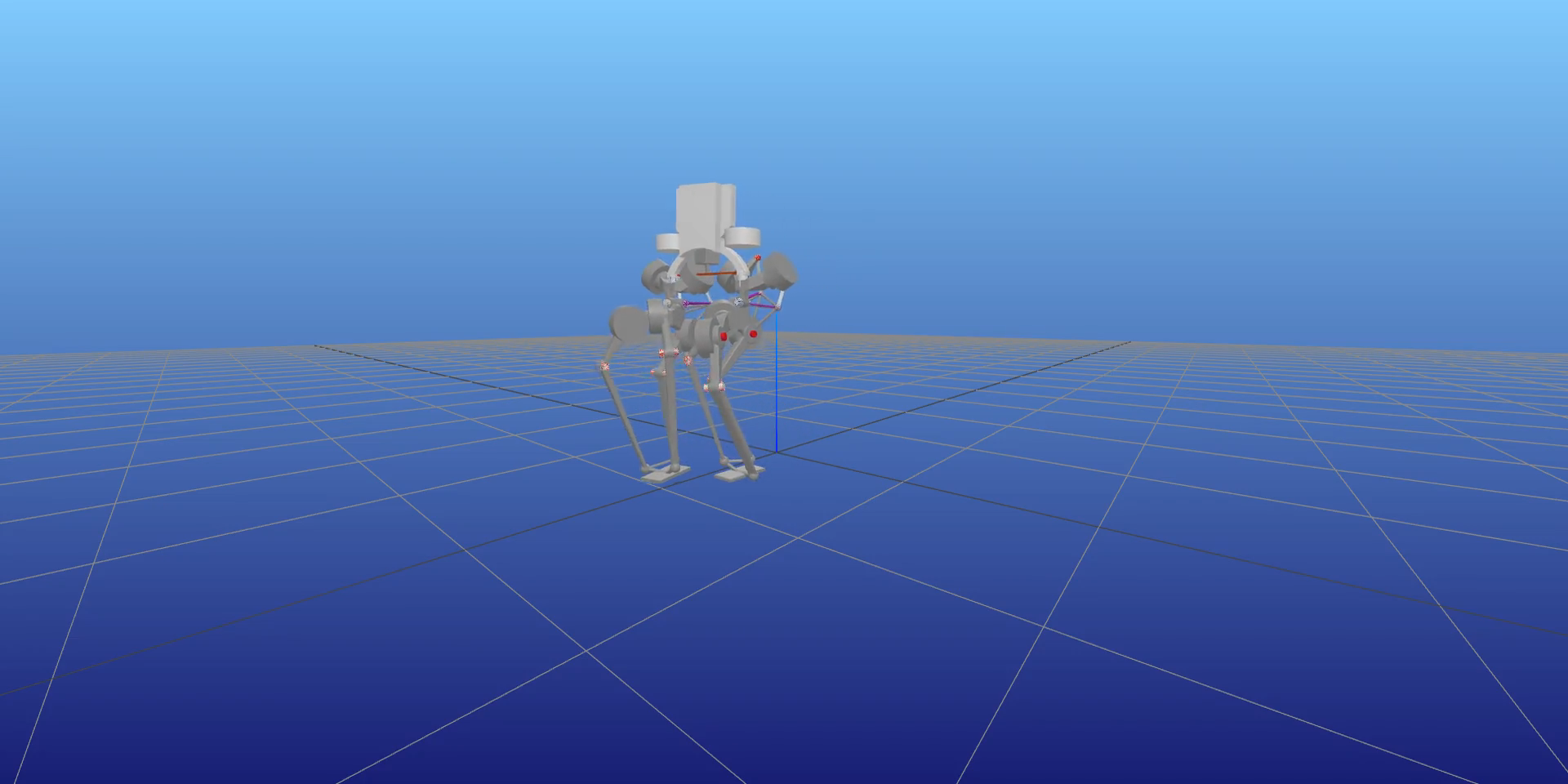}
        \caption{Walking}
    \end{subfigure}
    \hfill
    \begin{subfigure}{0.45\linewidth}
        \centering
        \includegraphics[trim={18cm 8cm 18cm 3cm}, clip, width=\linewidth]{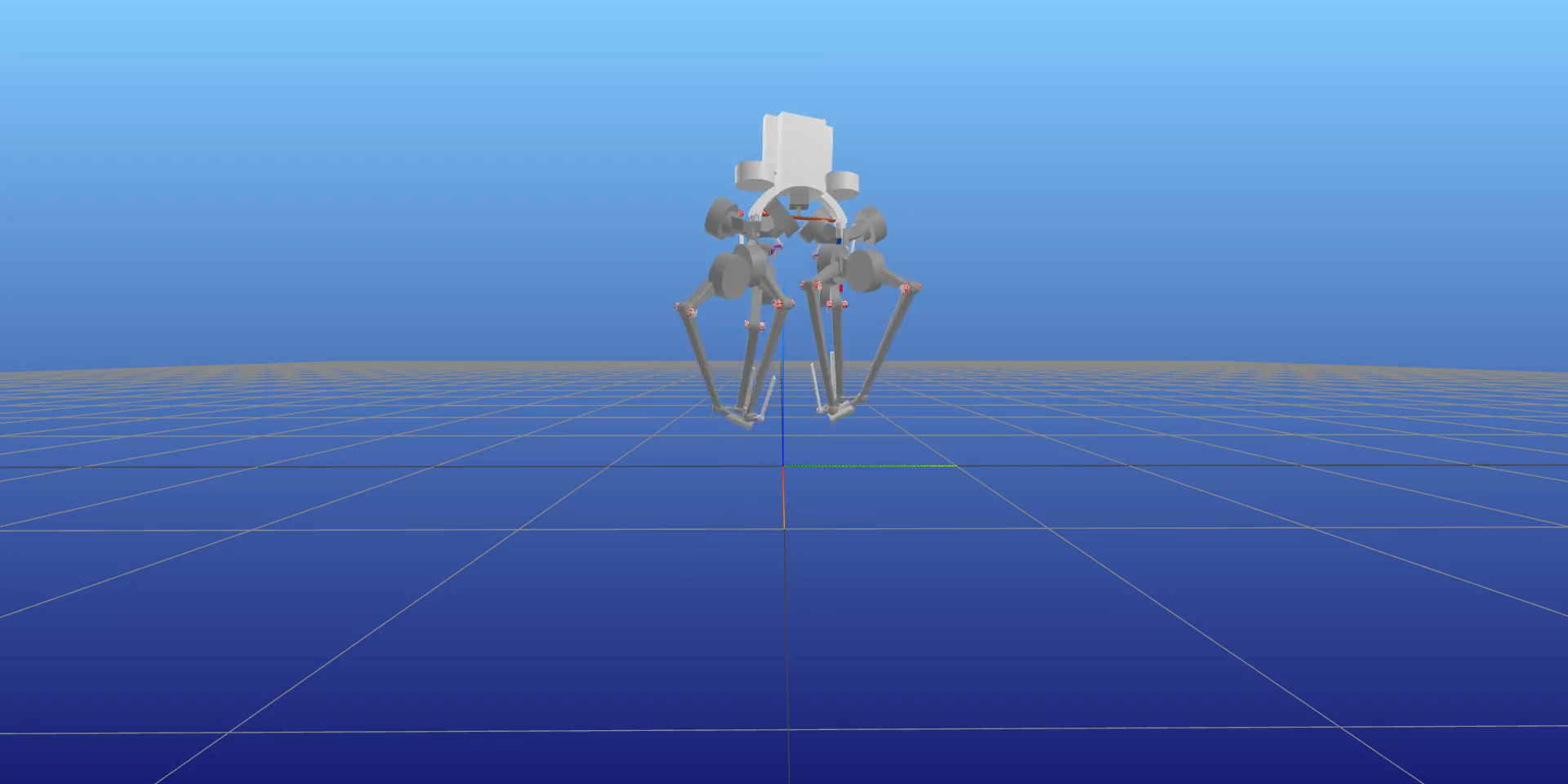}
        \caption{Jumping}
    \end{subfigure}
    \hfill
    \caption{Snapshots of motions performed on the Cleobot walker \cite{batto:hal-04717159}}
    \label{fig:snapshots_cleobot}
\end{figure}

Using the same costs as before, we perform squat, walk and jump motions on this architecture, as presented in Fig.~\ref{fig:snapshots_cleobot}, validating the generality of the approach.
Examples of such motions can be found in the companion video and reproduced using the code provided \cite{GithubClosedLoopMotion}.
It would not be possible to generate similar movements for this robot without our method since no approximate serial model is available and adjusting an equivalent serial kinematics would be very difficult. 
This result shows that our method enables the design of more complex walking kinematics, opening a possible research direction toward more effective walkers.

\subsection{Solver timings}
Optimizing a walking trajectory consisting in $T=250$ time-steps of $\Delta t = 15ms$ - i.e. a $3.75$s trajectory - takes about 500ms per iteration using our implementation in Crocoddyl \cite{carlosmastalliCrocoddylEfficientVersatile2019} with an 8 core parallelization.
In this time, 43\% is spend in the rollout of the problem - i.e. the forward dynamics, 30\% in the dynamics derivatives and 25\% in the backward pass of the FDDP algorithm.
We note that, as expected, reducing the number of closed-loop constraints reduces the computation time for both the forward dynamics and its derivatives. 
This results shows that our additional derivatives computation induces minimal loss in computational efficiency, opening the way for real time implementation.

\section{Conclusion} \label{sec:conclusion}
This paper presents a method to account for closed-loop kinematics in locomotion.
Our method relies on establishing an efficient differentiable model of the closed-loop constraints and implementing it in an optimal control solver.
We propose a complete implementation based on the open-source solver Crocoddyl. 
To validate our method, we developed a benchmark for comparing the motions obtained using an \textit{Approximate Serial} model, that relies on an underlying serial chain and ignores the closed-loop transmission, to motions obtained with the \textit{Closed-Kinematics} model.
We showed the limitations of the \textit{Approximate Serial} model on squats, walk, and stair climbing motions, revealing issues in joint torques estimation and actuator limits.
We then proved the generality of our method by generating motions on a robot for which an \textit{Approximate Serial} model cannot be defined, overcoming the limitations of previous methods.
% Future work will focus on applying and benchma refining the \textit{Approximate Serial} model to take into consideration the actuator specificity when it is possible, and on generalizing the method to more tasks and robots.
Our method directly improves the efficiency and dynamic range of actual robots and paves the road to more sophisticated actuation.

\bibliographystyle{ieeetr}
\bibliography{references}

\begin{thebibliography}{10}

\bibitem{tadeuszmikolajczykRecentAdvancesBipedal2022}
T.~Mikolajczyk, E.~Miko{\l}ajewska, H.~F. Al-Shuka, T.~Malinowski, A.~K{\l}odowski, D.~Y. Pimenov, T.~Paczkowski, F.~Hu, K.~Giasin, D.~Miko{\l}ajewski, {\em et~al.}, ``Recent advances in bipedal walking robots: Review of gait, drive, sensors and control systems,'' {\em Sensors}, vol.~22, no.~12, p.~4440, 2022.

\bibitem{jean-pierremerletParallelRobotsSolid2006}
J.-P. Merlet, {\em Parallel robots}, vol.~128.
\newblock Springer Science \& Business Media, 2006.

\bibitem{virgilebattoComparativeMetricsAdvanced2023}
{Virgile Batto}, {T. Flayols}, {N. Mansard}, and {Margot Vulliez}, ``Comparative {{Metrics}} of {{Advanced Serial}}/{{Parallel Biped Design}} and {{Characterization}} of the {{Main Contemporary Architectures}},'' {\em IEEE-RAS International Conference on Humanoid Robots}, 2023.

\bibitem{carlosmastalliCrocoddylEfficientVersatile2019}
C.~Mastalli, R.~Budhiraja, W.~Merkt, G.~Saurel, B.~Hammoud, M.~Naveau, J.~Carpentier, L.~Righetti, S.~Vijayakumar, and N.~Mansard, ``Crocoddyl: An efficient and versatile framework for multi-contact optimal control,'' {\em IEEE International Conference on Robotics and Automation (ICRA)}, 2020.

\bibitem{g.f.liuAnalysisControlRedundant2001}
G.~Liu, Y.~Wu, X.~Wu, Y.~Kuen, and Z.~Li, ``Analysis and control of redundant parallel manipulators,'' {\em IEEE International Conference on Robotics and Automation (ICRA)}, vol.~4, pp.~3748--3754, 2001.

\bibitem{FourierGR1}
``{{Fourier Intelligence}}.''
\newblock Accessed on 2024-09-12.

\bibitem{HumanoidRobotG1_Humanoid}
``Humanoid robot {{G1}} | {{Unitree Robotics}}.''
\newblock Accessed: 2024-09-12.

\bibitem{TeslaOptimus}
``{{Tesla AI \& robotics}}.''
\newblock Accessed on 2024-09-12.

\bibitem{AdamPnd}
``{{PNDbotics}}.''
\newblock Accessed on 2024-09-12.

\bibitem{AgilityProducts}
``{{AgilityProducts}}.''
\newblock Accessed on 2024-09-12.

\bibitem{roigHardwareDesignControl2022a}
A.~Roig, S.~K. Kothakota, N.~Miguel, P.~Fernbach, E.~M. Hoffman, and L.~Marchionni, ``On the {{Hardware Design}} and {{Control Architecture}} of the {{Humanoid Robot Kangaroo}},'' in {\em 6th {{Workshop}} on {{Legged Robots}} during the {{International Conference}} on {{Robotics}} and {{Automation}} ({{ICRA}} 2022)}, 2022.

\bibitem{keving.gimDesignFabricationBipedal2018}
K.~G. Gim, J.~Kim, and K.~Yamane, ``Design and fabrication of a bipedal robot using serial-parallel hybrid leg mechanism,'' in {\em IEEE/RSJ International Conference on Intelligent Robots and Systems (IROS)}, 2018.

\bibitem{hubicki2016atrias}
C.~Hubicki, J.~Grimes, M.~Jones, D.~Renjewski, A.~Spr{\"o}witz, A.~Abate, and J.~Hurst, ``Atrias: Design and validation of a tether-free 3d-capable spring-mass bipedal robot,'' {\em The International Journal of Robotics Research}, vol.~35, no.~12, pp.~1497--1521, 2016.

\bibitem{boukheddimi2023investigations}
M.~Boukheddimi, R.~Kumar, S.~Kumar, J.~Carpentier, and F.~Kirchner, ``Investigations into exploiting the full capabilities of a series-parallel hybrid humanoid using whole body trajectory optimization,'' in {\em 2023 IEEE/RSJ International Conference on Intelligent Robots and Systems (IROS)}, pp.~10433--10439, IEEE, 2023.

\bibitem{peekema2015template}
A.~T. Peekema, ``Template-based control of the bipedal robot atrias,'' 2015.

\bibitem{ramezani2014performance}
A.~Ramezani, J.~W. Hurst, K.~Akbari~Hamed, and J.~W. Grizzle, ``Performance analysis and feedback control of atrias, a three-dimensional bipedal robot,'' {\em Journal of Dynamic Systems, Measurement, and Control}, vol.~136, no.~2, p.~021012, 2014.

\bibitem{hongkaidaiWholebodyMotionPlanning2014}
H.~Dai, A.~Valenzuela, and R.~Tedrake, ``Whole-body motion planning with centroidal dynamics and full kinematics,'' {\em IEEE-RAS International Conference on Humanoid Robots}, pp.~295--302, 2014.

\bibitem{rawlings2017model}
J.~B. Rawlings, D.~Q. Mayne, M.~Diehl, {\em et~al.}, {\em Model predictive control: theory, computation, and design}, vol.~2.
\newblock Nob Hill Publishing Madison, WI, 2017.

\bibitem{ewendantecFirstOrderApproximation2022}
E.~Dantec, M.~Taix, and N.~Mansard, ``First order approximation of model predictive control solutions for high frequency feedback,'' {\em IEEE Robotics and Automation Letters (RA-L)}, vol.~7, no.~2, pp.~4448--4455, 2022.

\bibitem{elliotchane-saneCaTConstraintsTerminations2024}
E.~Chane-Sane, P.-A. Leziart, T.~Flayols, O.~Stasse, P.~Sou{\`e}res, and N.~Mansard, ``Cat: Constraints as terminations for legged locomotion reinforcement learning,'' {\em IEEE/RJS International Conference on Intelligent RObots and Systems}, 2024.

\bibitem{guillermoa.castilloRobustFeedbackMotion2021}
{Guillermo A. Castillo}, {Bowen Weng}, {Wei Zhang}, and {Ayonga Hereid}, ``Robust {{Feedback Motion Policy Design Using Reinforcement Learning}} on a {{3D Digit Bipedal Robot}},'' {\em IEEE/RJS International Conference on Intelligent Robots and Systems}, 2021.

\bibitem{royfeatherstoneRigidBodyDynamics2007}
R.~Featherstone, {\em Rigid body dynamics algorithms}.
\newblock Springer, 2014.

\bibitem{justincarpentierPinocchioLibraryFast2019}
J.~Carpentier, G.~Saurel, G.~Buondonno, J.~Mirabel, F.~Lamiraux, O.~Stasse, and N.~Mansard, ``The {Pinocchio} {C++} library: A fast and flexible implementation of rigid body dynamics algorithms and their analytical derivatives,'' {\em IEEE/SICE International Symposium on System Integration (SII)}, pp.~614--619, 2019.

\bibitem{martinl.felisRBDLEfficientRigidbody2016}
M.~L. Felis, ``{RBDL}: an efficient rigid-body dynamics library using recursive algorithms,'' {\em Autonomous Robots}, vol.~41, no.~2, pp.~495--511, 2017.

\bibitem{justincarpentierProximalSparseResolution2021}
J.~Carpentier, R.~Budhiraja, and N.~Mansard, ``Proximal and sparse resolution of constrained dynamic equations,'' {\em Robotics: Science and Systems}, 2021.

\bibitem{shubhamsinghAnalyticalSecondOrderDerivatives2023}
{Shubham Singh}, {Ryan P. Russell}, and {Patrick M. Wensing}, ``Analytical {{Second-Order Derivatives}} of {{Rigid-Body Contact Dynamics}}: {{Application}} to {{Multi-Shooting DDP}},'' {\em IEEE-RAS International Conference on Humanoid Robots}, 2023.

\bibitem{diehl:inria-00390435}
M.~Diehl, H.~G. Bock, H.~Diedam, and P.-B. Wieber, ``{Fast Direct Multiple Shooting Algorithms for Optimal Robot Control},'' {\em {Fast Motions in Biomechanics and Robotics}}, 2005.

\bibitem{stephaneredonGaussLeastConstraints2002}
S.~Redon, A.~Kheddar, and S.~Coquillart, ``Gauss' least constraints principle and rigid body simulations,'' {\em IEEE international conference on robotics and automation (ICRA)}, 2002.

\bibitem{davidbaraffFastContactForce1994}
D.~Baraff, ``Fast contact force computation for nonpenetrating rigid bodies,'' {\em International Conference on Computer Graphics and Interactive Techniques}, pp.~23--34, 1994.

\bibitem{firdause.udwadiaEquationsMotionMechanical1996}
F.~E. Udwadia, ``Equations of motion for mechanical systems: A unified approach.,'' {\em International Journal of Non-linear Mechanics}, vol.~31, no.~6, pp.~951--958, 1996.

\bibitem{justincarpentierAnalyticalDerivativesRigid2018}
J.~Carpentier and N.~Mansard, ``Analytical derivatives of rigid body dynamics algorithms,'' in {\em Robotics: Science and systems (RSS)}, 2018.

\bibitem{joansolaMicroLieTheory2018}
J.~Sola, J.~Deray, and D.~Atchuthan, ``A micro lie theory for state estimation in robotics,'' {\em arXiv:1812.01537}, 2018.

\bibitem{sebastienkleffDerivationContactDynamics}
S.~Kleff, J.~Carpentier, N.~Mansard, and L.~Righetti, ``On the derivation of the contact dynamics in arbitrary frames: Application to polishing with talos,'' {\em IEEE-RAS 21st International Conference on Humanoid Robots (Humanoids)}, 2022.

\bibitem{Mronga2022WBC}
D.~Mronga, S.~Kumar, and F.~Kirchner, ``Whole-body control of series-parallel hybrid robots,'' 02 2022.

\bibitem{GithubParallelRobots}
``Example parallel robots: a repository containing several urdf models of legged robots with parallel kinematics - on github.''

\bibitem{batto:hal-04717159}
V.~Batto, L.~de~Matteis, T.~Flayols, M.~Vulliez, and N.~Mansard, ``{CLEO: Closed-Loop kinematics Evolutionary Optimization of bipedal structures}.'' working paper or preprint, Oct. 2024.

\bibitem{GithubCrocoddylFork}
``Github - crocoddyl - temporary fork.''

\bibitem{wilsonjalletPROXDDPProximalConstrained}
W.~Jallet, A.~Bambade, E.~Arlaud, S.~El-Kazdadi, N.~Mansard, and J.~Carpentier, ``Proxddp: Proximal constrained trajectory optimization,'' 2023.

\bibitem{armandjordanaStagewiseImplementationsSequential}
A.~Jordana, S.~Kleff, A.~Meduri, J.~Carpentier, N.~Mansard, and L.~Righetti, ``Stagewise implementations of sequential quadratic programming for model-predictive control,'' {\em Preprint}, 2023.

\bibitem{GithubClosedLoopMotion}
``Locomotion generation for parallel robots: a collection of ocp problems used to generate the examples of the paper - on github.''

\end{thebibliography}

\appendices
\section{From Approximate Serial model to Closed-Kinematics} \label{app:lifting}
As the leg is fully actuated and outside of singularities of the parallel linkage, every state trajectory of the \textit{Approximate Serial} model can be casted to a trajectory of the \textit{Closed-Kinematics} model with same serial state.
A trajectory for the serial model is defined by a sequence of states $x_s^{[k]} = (q_s^{*[k]}, \dot{q}_s^{*[k]})$ and of controls $u_s^{[k]}$.
A trajectory for the closed model is similarly defined by the sequences $x_c^{[k]} = (q_c^{[k]}, \dot{q}_c^{[k]})$ and $u_c^{[k]}$, where $q_c^{[k]} = \begin{pmatrix}
    q_s^{[k]} &
    q_l^{[k]}
\end{pmatrix}$ (respectively $\dot{q}_c$), describing that the \textit{Closed-Kinematics} model state includes the \textit{Approximate Serial} model state.
We propose to lift the \textit{Approximate Serial trajectory} into a \textit{Closed-Kinematics trajectory} by solving:
\begin{equation} \label{equ:open_to_closed_problem}
    \begin{aligned}
        \min_{u^{[k]}, x_c^{[k+1]}} & \quad \frac{1}{2} \norm{q_s^{[k+1]} - q_s^{*[k+1]}}^2 + \frac{1}{2} \norm{\dot{q}_s^{[k+1]} - \dot{q}_s^{*[k]}}^2\\
        \text{s.t.} & \quad x_c^{[k+1]} = f_k(x_c^{[k]}, u^{[k]}) \\
    \end{aligned}
\end{equation}
where the state $x_c^{[k]}$ is known from the previous iteration (assuming $x^{[0]}$ is known) and the targets are the expected values for the serial part of the state, given by the \textit{Approximate Serial trajectory}.
We can observe that problem \eqref{equ:open_to_closed_problem} takes the form of a 1-step optimal control problem and can therefore be solved using the same solver as before.

\end{document}